\documentclass[11pt, a4paper, copyright, nonumbering]{selftok}

\usepackage{dblfloatfix}
\usepackage{ulem}
\usepackage{caption}
\usepackage{dramatist}
\usepackage{xspace}
\usepackage{pifont} %
\usepackage{multirow}
\usepackage{tcolorbox}
\usepackage{xltabular}
\usepackage{longtable}
\usepackage{hyperref}
\usepackage{amsfonts}
\usepackage{amsmath}
\usepackage{amssymb}
\usepackage{lineno}
\usepackage{multirow}
\usepackage{adjustbox}
\usepackage{listings}
\usepackage{algorithm}
\usepackage{booktabs}
\usepackage{algpseudocode}
\usepackage{amsmath}

\usepackage{ulem}
\usepackage{ragged2e} 

\usepackage[bottom]{footmisc}

\usepackage{CJKutf8}
\usepackage{setspace}

\usepackage{dsfont}
\usepackage{array} %
\usepackage{tabularx} %
\usepackage{xcolor} %
\usepackage{tabularx}
\usepackage{booktabs}

\usepackage{lipsum}  %
\usepackage{multicol} %

\usepackage{caption}    
\usepackage{subcaption} 
\usepackage{wrapfig}

\newcommand{\ie}{\textit{i.e.}}
\newcommand{\eg}{\textit{e.g.}}

\newcommand{\ceil}[1]{\left\lceil #1 \right\rceil}

\makeatletter
\def\@BTrule[#1]{%
  \ifx\longtable\undefined
    \let\@BTswitch\@BTnormal
  \else\ifx\hline\LT@hline
    \nobreak
    \let\@BTswitch\@BLTrule
  \else
     \let\@BTswitch\@BTnormal
  \fi\fi
  \global\@thisrulewidth=#1\relax
  \ifnum\@thisruleclass=\tw@\vskip\@aboverulesep\else
  \ifnum\@lastruleclass=\z@\vskip\@aboverulesep\else
  \ifnum\@lastruleclass=\@ne\vskip\doublerulesep\fi\fi\fi
  \@BTswitch}
\makeatother

\addto\extrasenglish{
}

 {\begin{list}{}%
         {\setlength{\leftmargin}{#1}}%
         \item[]%
 }
 {\end{list}}

\reportnumber{001} %

\renewcommand{\today}{}

\title{\centering Selftok: Discrete Visual Tokens of Autoregression, by Diffusion, and for Reasoning}

\author[*]{
Selftok Team, Media Technology Institute, Huawei
}

\renewcommand{\phi}{\varphi}

\renewcommand{\geq}{\geqslant}

\renewcommand{\epsilon}{\varepsilon}
\renewcommand{\imath}{\mathrm{i}}

\newlength{\restsubwidth}
\newlength{\restsubheight}
\newlength{\restsubmoreheight}
\newcommand{\rest}[2]{%
        \settowidth{\restsubwidth}{\ensuremath{#2}}
        \settoheight{\restsubheight}{\ensuremath{{}_{#2}}}
        \ensuremath{{#1\hskip 0.5pt}_{\vrule\kern2pt\parbox[b][%
        4pt][b]{\the\restsubwidth}{%
                        \ensuremath{{}_{#2}}}}}
        }

\begin{abstract}
We completely discard the conventional \textit{spatial prior} in image representation and introduce a novel \textit{discrete} visual tokenizer: Self-consistency Tokenizer (Selftok). At its design core, we compose an autoregressive (AR) prior---mirroring the causal structure of language---into visual tokens by using the reverse diffusion process of image generation. The \textit{AR property} makes Selftok fundamentally distinct from traditional spatial tokens in the following two key ways:

\vspace{1mm}
$\bullet$  Selftok offers an elegant and minimalist approach to \textit{unify diffusion and AR} for vision-language models (VLMs): By representing images with Selftok tokens, we can train a VLM using a purely discrete autoregressive architecture---like that in LLMs---without requiring additional modules or training objectives. \\
$\bullet$ We theoretically show that \textit{the AR prior satisfies the Bellman equation}, whereas the spatial prior does not. Therefore, Selftok supports reinforcement learning (RL) for visual generation with effectiveness comparable to that achieved in LLMs.
\vspace{1mm}

Besides the AR property, Selftok is also a  SoTA tokenizer that achieves a favorable trade-off between high-quality reconstruction and compression rate. We use Selftok to build a pure AR VLM for both visual comprehension and generation tasks. Impressively, without using any text-image training pairs, a simple policy gradient RL working in the visual tokens can significantly boost the visual generation benchmark, surpassing all the existing models by a large margin. Therefore, we believe that Selftok effectively addresses the long-standing challenge that visual tokens cannot support effective RL. When combined with the well-established strengths of RL in LLMs, this brings us one step closer to realizing a truly multimodal LLM. Project Page: \url{https://selftok-team.github.io/report/}.

\end{abstract}

\begin{document}

\maketitle

\begin{figure*}[!h]
    \centering
    \footnotesize
         \includegraphics[width=.9\textwidth]{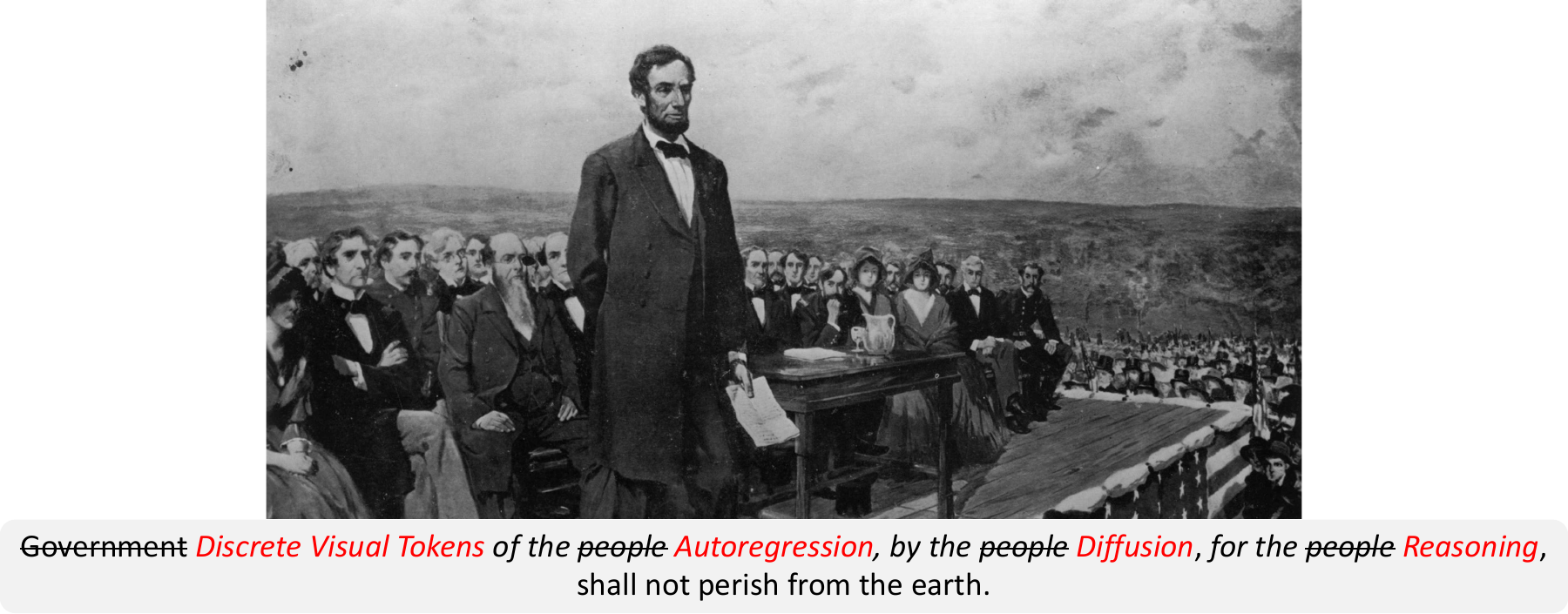}
         \phantomcaption
    \vspace*{-5mm}
\end{figure*}

\newpage

\begin{spacing}{0.9}
\tableofcontents
\end{spacing}

\newpage

\vspace{-3mm}
\section{Introduction}
\label{sec:1}
\vspace{-2mm}
There is a growing consensus that language data will soon be exhausted by large language models (LLMs), whereas non-language data like images or videos are significantly underutilized~\cite{sutskever2024peakdata}. Therefore, unifying multiple modalities is a key step toward unlocking AI's more powerful emergent capabilities. To unify all modalities into one discrete autoregressive model (dAR) like LLMs, we need to tokenize non-language modalities into \textit{discrete} tokens\footnote[1]{Language also needs tokenization like BPE~\cite{sennrich2016bpe}. It is easier than that for images and hence outside our scope.}. In this paper, we only discuss \textit{visual} discrete tokens and use them together with language to train a vision-language model (VLM) that can do both visual comprehension (\ie, language output like VQA) and generation (\ie, visual output like image synthesis).  

\begin{figure*}
    \centering
    \footnotesize
    \begin{subfigure}[t]{\textwidth}
         \includegraphics[width=\textwidth]{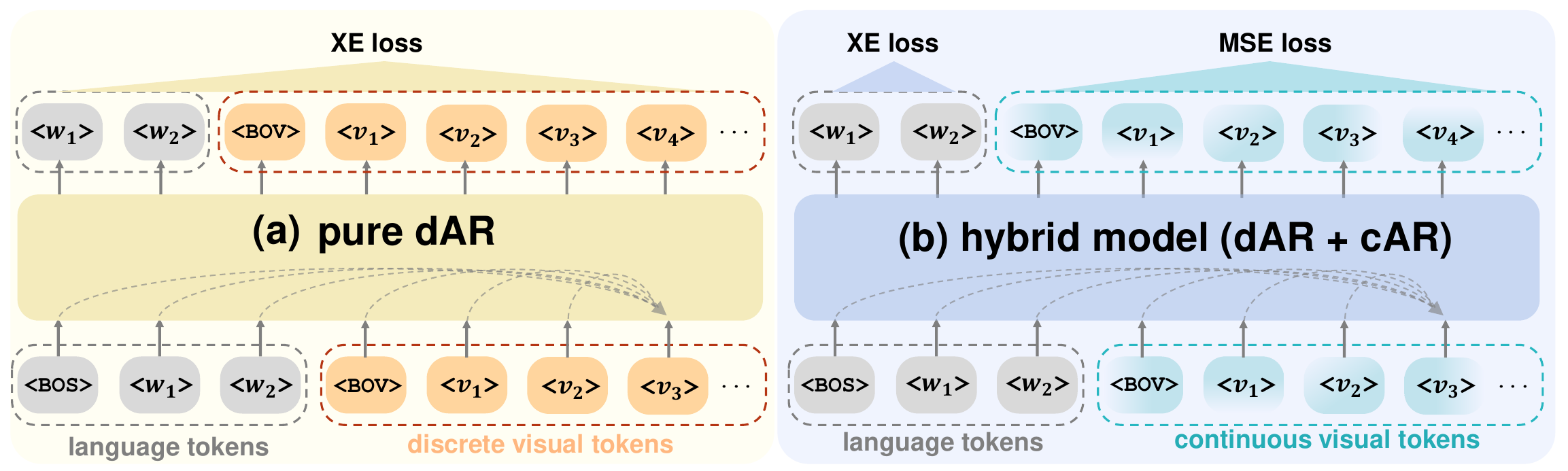}
         \phantomcaption
         \label{fig:1a}
    \end{subfigure}
    \begin{subfigure}[t]{0\textwidth} 
         \includegraphics[width=0\textwidth]{example-image-b}
         \phantomcaption
         \label{fig:1b}   
    \end{subfigure}
    \vspace*{-9mm}
    \caption{Comparison of (a) pure discrete autoregressive model (dAR) and (b) hybrid model that combines dAR and continuous autoregressive model (cAR). \texttt{<BOS>}/\texttt{<BOV>} indicates the start of a sentence/image. \texttt{<$w_i$>}/\texttt{<$v_i$>} denotes the $i$-th language/visual token. Both models predict the next token given all previous ones, \eg, [\texttt{<BOS>}, ..., \texttt{<$v_3$>}] $\to$\texttt{<$v_4$>}.}
    \vspace*{-1mm}
    \label{fig:1}
\end{figure*}
\vspace{-3mm}
\subsection{Why Discrete?}
\label{sec:1.1}
Before introducing our Selftok visual tokenizer, we need to clarify why we advocate the use of a pure dAR (Figure~\ref{fig:1a}), rather than a hybrid approach that combines a dAR for language and a continuous autoregressive model (cAR) for images (Figure~\ref{fig:1b})~\cite{zhou2024transfusion,li2024autoregressive}. The latter is widely adopted by proponents who argue that visual data should be encoded as continuous tokens to minimize the compression loss, but this is just a minor concern---there are many post-processing methods available to ensure the precision~\cite{duggal2024adaptive,miwa2025one,tanghart,li2024imagefolder}. However, using cAR (or hybrid) leads to major issues that cannot be fundamentally resolved without adopting a pure dAR:
\vspace{-2mm}
\begin{itemize}[noitemsep, left=0pt]
    \item \textbf{cAR cannot inherit the successful infrastructure and training paradigm of LLMs.} This is the most common reason cited in existing dAR-based VLMs~\cite{team2024chameleon, wang2024emu3, agarwal2025cosmos, chen2025janus, kondratyuk2023videopoet}. Yet, the following three justifications are often overlooked by the community.
    
    \item \textbf{cAR is more error-prone in next-token prediction}. While dAR functions as a sequential token classifier trained with cross-entropy (XE) loss, cAR operates as a sequential vector regressor trained with mean squared error (MSE) loss, which is less stable and harder to optimize than XE~\cite{hu2022understanding,chen2024next}. Perhaps this is the key reason why most cARs abandon the causal next-token prediction and revert to bidirectional modeling, such as demasking~\cite{li2024autoregressive,yu2024randomized} or holistic reconstruction~\cite{zhou2024transfusion, ma2024janusflow}. Unfortunately, they undermine the core design philosophy of the decoder-only AR: the causal dependency of tokens~\cite{radford2018improving}.
    
    \item \textbf{cAR introduces unnecessary complexity into reinforcement learning (RL)}. It is widely known that RL is an indispensable post-training step to unleash the power of LLMs~\cite{guo2025deepseek}. However, cAR turns the finite Markov Decision Process (MDP) formulation of dAR---with a discrete state-action space---into an infinite MDP with a continuous state-action space, thereby complicating policy optimization~\cite{meyer2023harnessing}.
    \item \textbf{Continuous representations are less disentangled than discrete ones}. Disentanglement uncovers the modular and true generative factors of data~\cite{higgins2018towards,wang2021self}, which are critical for: 1) Unbiased visual comprehension, \eg, if ``color'' and ``object'' are disentangled, the model can still recognize a \textit{black swan} as \textit{swan}, even if all the training examples of \textit{swans} are \textit{white}; and 2) Controlled generation, if such disentanglement holds, the model can generate a \textit{black swan} without seeing one in training. 
    Since a real-valued vector is infinitely countable, a single continuous token may theoretically entangle all the factor combinations. As a result, achieving disentanglement would require an impractically large amount of training data to cover all the combinations~\cite{locatello2019challenging}, \eg, we need $\mathcal{O}(N^M)$ images, where $N$ is \#values per factor and $M$ is the \#factors per image. In contrast, discrete tokens, with their limited information bandwidth, serve as a strong inductive bias that encourages disentanglement~\cite{hsu2023disentanglement}.
    \end{itemize}

\begin{figure*}
    \centering
    \footnotesize
    \includegraphics[width=0.95\textwidth]{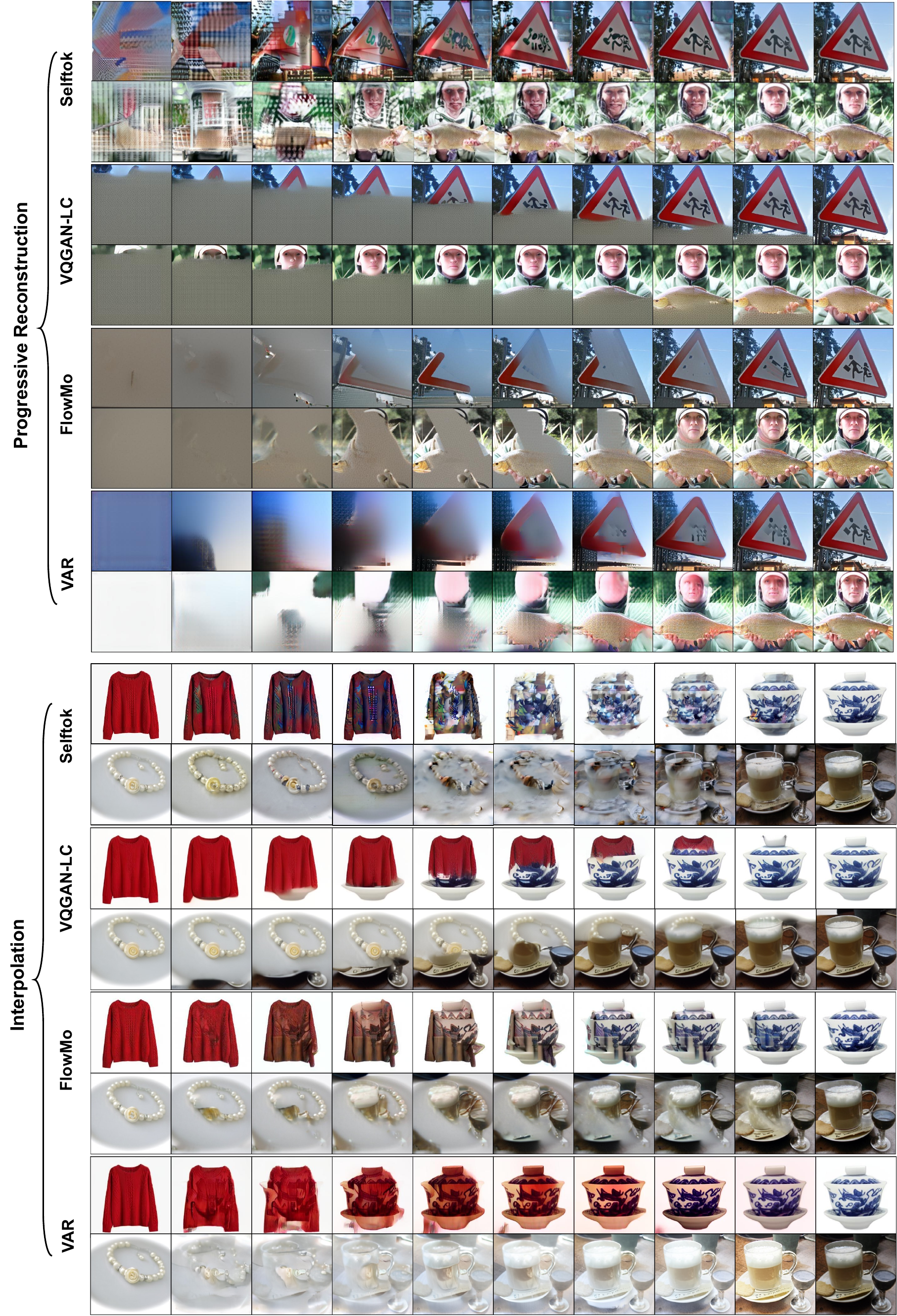}
    \vspace*{-2mm}
    \caption{Progressive reconstruction (left to right): Reconstructions by progressively masking out a shorter sequence of tokens before inputting to the decoder.
    Interpolation (left to right): Reconstructions by gradually replacing tokens of the left image with those of the right one.
    All methods except Selftok exhibit strong spatial characteristics (\ie, tokens$\Leftrightarrow$patches).}
    \vspace*{-1mm}
    \label{fig:disentangle}
\end{figure*}

\subsection{Why Not Spatial?}
\label{sec:1.2}

Since representing an image as a set of spatial grids has been the standard approach back in the early days of computer vision, it is often assumed---almost by default---that spatial tokens are naturally suited for AR modeling like language. However, we overlook the fact that the \textit{causal} dependencies among \textbf{spatial tokens are inherently not autoregressive} and thus cannot be modeled in the same way as language using AR models. 

\begin{figure*}
    \centering
    \footnotesize
    \begin{subfigure}[t]{\textwidth}
         \includegraphics[width=\textwidth]{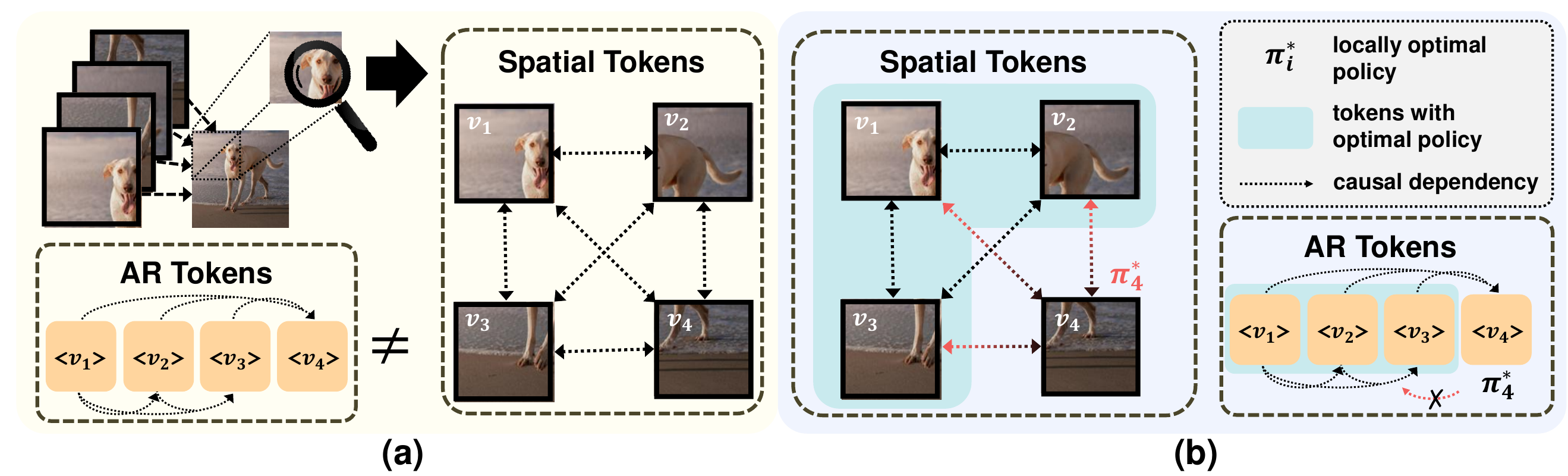}
         \phantomcaption
         \label{fig:2a}
    \end{subfigure}
    \begin{subfigure}[t]{0\textwidth} 
         \includegraphics[width=0\textwidth]{example-image-b}
         \phantomcaption
         \label{fig:2b}   
    \end{subfigure}
    \vspace*{-11mm}
    \caption{(a) As an image can be viewed as the effect of spatial pixels (or patches), observing any part of it introduces spurious dependencies among spatial tokens, making them non-AR. (b) Due to the anti-causal links (red) for spatial tokens, learning the locally optimal policy $\pi^*_4$ for a later token (\eg, $v_4$) can propagate backward and interfere with earlier tokens that were already optimized (\eg, $v_1,v_2,v_3$). In contrast, AR tokens without such links do not have this issue. A more formal illustration is in Section~\ref{sec:4.1}.}
    \vspace*{-3mm}
    \label{fig:2}
\end{figure*}

First, as shown in Figure~\ref{fig:2a}, since spatial pixels (the cause) collectively form the image (the effect), observing any part of the image during spatial token encoding can introduce spurious dependencies among the tokens due to the collider effect~\cite{pearl2009causality}\footnote[2]{If the causal graph is $A\to C\leftarrow B$, $C$ is the collider; even if $A$ and $B$ are independent, observing $C$ may cause dependencies between $A$ and $B$.}, \ie, the resulting Bayesian causal graph of spatial tokens is not as autoregressive as that of AR tokens (Figure~\ref{fig:entropy_token}).

Second, we reveal a significant drawback of spatial tokens that has previously gone unnoticed: \textbf{non-AR dependency fundamentally violates the policy improvement optimality} in RL~\cite{sutton1998reinforcement}. To illustrate this in Figure~\ref{fig:2b}, in non-AR dependency, token prediction (action) at a later time affects the tokens predicted in earlier steps (earlier states), so the later policy may contradict earlier policies that have already been optimized. Compared to AR tokens, RL for non-AR spatial tokens is expected to be significantly less effective (Section~\ref{sec:4.3}). 

Based on the above two reasons, we conjecture that the misuse of spatial tokens---whether discrete or continuous--- is the key reason why AR-based VLMs do not scale as effectively as LLMs. We leave the large-scale empirical validation of this hypothesis to future work.

\subsection{Contributions}
\label{sec:1.3}
We propose \textbf{Selftok: \emph{Self}-consistency \emph{Tok}enizer}, which encodes an image into \textit{autoregressive discrete} tokens, where each token corresponds to a diffusion time step. Therefore, Selftok can be integrated into a single dAR-based VLM, which can \textit{reason like LLMs but in the visual domain}. The following summarizes our contributions and provides a reading guide to the rest of the paper.

\noindent\textbf{1) Selftok \emph{of} Autoregression}: Compared to conventional spatial tokens, Selftok completely abandons the long-standing spatial prior without compromising reconstruction quality (Section~\ref{sec:2.3}) and spatial understanding (Section~\ref{sec:3.2}). More importantly, thanks to the AR property, Selftok offers better language compatibility of images, leading to improved dAR-based VLM training (Section~\ref{sec:3}) and effective RL-based post-training (Section~\ref{sec:4}).

\noindent\textbf{2) Selftok \emph{by} Diffusion}: Selftok leverages the AR nature of the reverse diffusion process and encodes the entire trajectory of image generation (Figure~\ref{fig:disentangle}\&Section~\ref{sec:2.1}). Therefore, it offers one of the most elegant yet straightforward approaches to unify diffusion and AR into a single dAR framework, in contrast to existing methods that rely on additional architectures or training objectives incompatible with dAR.

\noindent\textbf{3) Selftok \emph{for} Reasoning}: Thanks to its AR property, Selftok produces visual tokens that satisfy the optimality condition of the policy improvement (Section~\ref{sec:4}). As a result, Selftok-based VLMs support effective RL-based post-training for visual generation, akin to LLMs, whereas spatial token-based counterparts do not. For example, without using any pairwise supervision, our Selftok-Zero achieves impressive image generation performances on GenEval: 92\% (Table~\ref{tab:main_geneval}) and DPG-Bench: 85.57 (Table~\ref{tab:main_DPG}).

The preliminary core ideas of Selftok are presented in~\cite{yue2024exploring,pan2025ddt,lin2025phyvideo}. We encourage readers to begin with them before turning to this extended version, which offers a more theoretical analysis and presents more comprehensive and improved results.

\section{Selftok: Self-consistency Tokenizer}
\label{sec:2}

We begin by unpacking the meaning behind the name \textit{Self-consistency}. \textbf{``Self''} is just a rephrasing of the concept behind auto-encoder: both the encoder and decoder are trained to reconstruct the input image itself.

\begin{figure*}
    \centering
    \footnotesize
    \begin{subfigure}[t]{0.97\textwidth}
         \includegraphics[width=\textwidth]{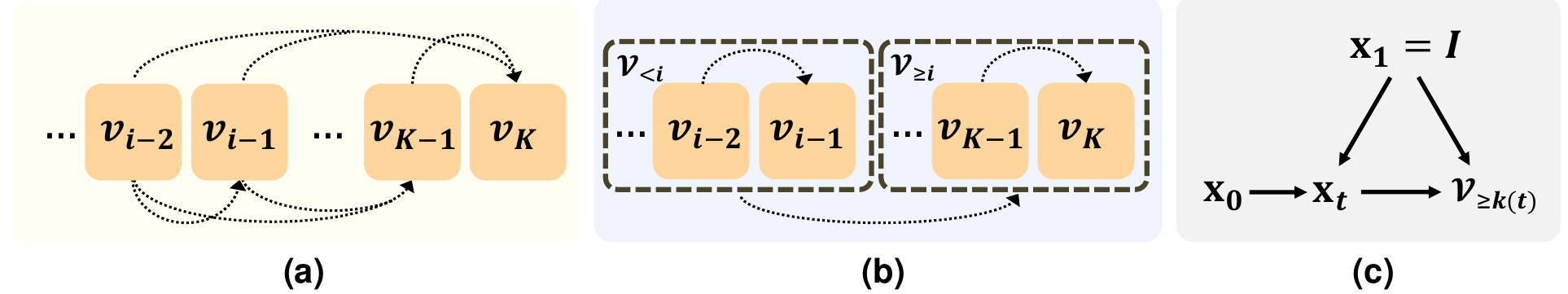}
         \phantomcaption
         \label{fig:4a}
    \end{subfigure}
    \begin{subfigure}[t]{0\textwidth} 
         \includegraphics[width=0\textwidth]{example-image-b}
         \phantomcaption
         \label{fig:4b}   
    \end{subfigure}
    \begin{subfigure}[t]{0\textwidth} 
         \includegraphics[width=0\textwidth]{example-image-b}
         \phantomcaption
         \label{fig:4c}
    \end{subfigure}

    \caption{(a) The causal graph for AR, where each dotted direct edge represents a causation. (b) The recursion of the AR causal graph. (c) The causal graph for learning $\mathcal{V}_{\geq k(t)}$ in Eq.~\eqref{eq:rec}.}

    \label{fig:4}
\end{figure*}

Specifically, our goal is to encode an image $I$ into $K$ discrete tokens, \ie, $\mathrm{Enc}(I) = \mathcal{V}_K = [v_1, v_2, ..., v_K]$, which can be decoded to reconstruct $I$ while adhering to an autoregressive (AR) prior. We formulate the following constrained optimization:
\begin{equation}
    \label{eq:main}
    \begin{aligned}    &\mathop{\mathrm{min}}_{\mathrm{Enc}(I) = \mathcal{V}_K,~\mathrm{Dec}} ~\|I - \mathrm{Dec}\left(\mathcal{V}_K\right)\|^2,\\
    s.t. \;\; P(\mathcal{V}_K) \overset{\text{AR}}{=} P&(v_1)\cdot P(v_2|v_1)\cdot \ldots \cdot P(v_K|v_1,\ldots, v_{K-1}),
    \end{aligned}
\end{equation}
where we define $\overset{\text{AR}}{=}$ as a special equality to indicate that the tokens $\mathcal{V}_K$ conform to the AR causal graph in Figure~\ref{fig:4a}, \ie, each token is generated from its predecessors\footnote[3]{This can be written mathematically as $P\left(\mathcal{V}_{< i} | do(\mathcal{V}_{\geq i})\right) = P(\mathcal{V}_{< i}) \;\forall i\in \{1,\ldots,K\}$ using the do-calculus~\cite{pearl2009causality}.}.
This definition is necessary, as the factorization is always valid for any token sequence through the chain rule of probability and does not necessarily imply an AR structure \textit{per se}.

As with other discrete compression problems~\cite{yang2023introduction}, solving the constrained optimization in Eq.~\eqref{eq:main} is inherently NP-hard due to the combinatorial nature of token assignment.
To make this tractable, we introduce an inductive bias grounded in the reverse diffusion process, which jointly satisfies the AR constraint and the reconstruction objective.
In particular, the term \textbf{``Consistency''} comes from Consistency Model~\cite{cm}. Similarly, we use a diffusion model and make the decoder consistent with the image generation path, \ie, reconstructing $\mathbf{x}_1 = I$ from any noisy inputs $\mathbf{x}_t$ along the path.

Next, in Section~\ref{sec:2.1}, we show how the reverse diffusion process can be formulated in a recursive fashion that enables efficient learning of AR tokens. Then, we present implementation details in Section~\ref{sec:2.2} and validation results in Section~\ref{sec:2.3}.

\subsection{Autoregression and Recursion by Diffusion}
\label{sec:2.1}
We show in Figure~\ref{fig:4b} that AR structure has an equivalent recursion, enabling a divide-and-conquer approach that decomposes the challenging constraint in Eq.~\eqref{eq:main} into simpler ones:
\begin{equation}
     P(\mathcal{V}_K) \overset{\text{AR}}{=} P(\mathcal{V}_{<i}) \cdot P(\mathcal{V}_{\geq i}|\mathcal{V}_{<i}),
     \label{eq:recursive}
 \end{equation}
where $\mathcal{V}_{<i} = [v_1, v_2, ..., v_{i-1}]$ and $\mathcal{V}_{\geq i} = [v_i, v_{i+1}, ..., v_{K}]$. For example, we can recursively apply Eq.~\eqref{eq:recursive} until it becomes a trivial learning problem $P(\mathcal{V}_{<K}) \cdot P(v_K|\mathcal{V}_{<K})$: if $\mathcal{V}_{<K}$ is provided, it is easy to encode the last token $v_K$.
Interestingly, the reverse diffusion process (in ODE form) has a similar decomposition~\cite{liuflow,songscore}:
\begin{equation}
  \frac{d\mathbf{x}_t}{dt}  = \mathbf{v}_t(\mathbf{x}_t),~~t\in[0,1]\;\;\;\; \overset{\text{solution}}{\Longrightarrow} \;\;\;\;\;\; \underbrace{\mathbf{x}_1}_{\text{destination}} =  \underbrace{\mathbf{x}_t}_{\text{midway  point }} + \;\;\underbrace{\int_{t}^1 \mathbf{v}_s(\mathbf{x}_s) ds}_{\substack{\textrm{path from midway } \\ \textrm{to destination:}~\mathbf{x}_t \rightsquigarrow \mathbf{x}_1}},
 \label{eq:ode}
\end{equation}
where $\mathbf{v}_t(\mathbf{x}_t)$ is the velocity field at time-step $t$ that transports the noisy midway $\mathbf{x}_t$, starting from $\mathbf{x}_0\in\mathcal{N}(0,1)$, towards the clean image $\mathbf{x}_1=I$. This shows that, if the midway $\mathbf{x}_t$ is provided, the reconstruction of $\mathbf{x}_1$ starting from $\mathbf{x}_t$ is easier than directly moving from $x_0$ to $x_1$.

Hence, we can establish a correspondence between the two recursions by aligning the provided midway point (part 1) and what comes after it (part 2), respectively:
\begin{equation}
 \left(\underbrace{ P(\mathcal{V}_K) \Longleftrightarrow 
 \mathbf{x}_1 }_{\text{Whole}} \right) = \left(\underbrace{ P(\mathcal{V}_{<i}) \Longleftrightarrow 
 \mathbf{x}_t }_{\text{Part 1}} \right)\;\; + \;\; \left(\underbrace{P(\mathcal{V}_{\geq i}|\mathcal{V}_{<i})\Longleftrightarrow\int_{t}^1 \mathbf{v}_s(\mathbf{x}_s) ds}_{\text{Part 2}} \right).
 \label{eq:correspondence}
\end{equation}
Motivated by this, we aim to compose the AR constraint into the reconstruction in Eq.~\eqref{eq:main}. Specifically, we decompose the entire reconstruction (from pure noise $\mathbf{x}_0$ to $\mathbf{x}_1$) into two parts with a similar recursion:
Part 1: A given $\mathbf{x}_t$, sampled from the diffusion path $q(\mathbf{x}_t|\mathbf{x}_1)$, encapsulates $\mathcal{V}_{<i}$, which is assumed to be already encoded; and Part 2: The reconstruction from $\mathbf{x}_t$ to $\mathbf{x}_1$ for learning the tokens $\mathcal{V}_{\geq i}=[v_i, v_{i+1},\ldots, v_{K}]$. Now, we present the Selftok training objective for an image sample $\mathbf{x}_1 = I$:
\begin{equation}
   \text{Selftok objective}: \mathop{\mathrm{min}}_{\substack{\mathrm{Enc}(\mathbf{x}_1) = \mathcal{V}_K, \\ \mathrm{Dec}}} ~~\mathop{\mathbb{E}}_{t\in [0,1]} \;\left[ \mathop{\mathbb{E}}_{\mathbf{x}_t\sim q(\mathbf{x}_t|\mathbf{x}_1)} \left[ \| \mathbf{x}_1 - \mathrm{Dec}(\mathbf{x}_t, \mathcal{V}_{\geq k(t)}) \|^2 \right] \right],
    \label{eq:rec}
\end{equation}
where $\mathcal{V}_{\geq k(t)} = [v_{k(t)}, v_{k(t)+1}, ..., v_{K}]$ and $k(t)$ is a token schedule with $k(1) = K+1$ and $k(0)=1$, which maps each continuous time-step $t$ to a discrete token index $i$ in Eq.~\eqref{eq:correspondence}. The choices of $q(\mathbf{x}_t|\mathbf{x}_1)$ and $k(t)$ are discussed in Section~\ref{sec:2.2.3}\&\ref{sec:2.2.4}, respectively. When the context is clear, we use $k(t)$ and $i$ interchangeably.  We highlight that our Sefltok is indeed \textbf{non-spatial}: $\mathcal{V}_K$ discretizes the continuous velocity field of the entire image generation path, which is beyond the na\"{i}ve spatial visual cues. Please recall Figure~\ref{fig:disentangle} for qualitative illustrations.

Here, we verify that the Seltok objective in Eq.~\eqref{eq:rec} optimizes the original one in Eq.~\eqref{eq:main} from the following three aspects: 
\\
1) \textbf{Reconstruction}: When $t = 0$, Eq.~\eqref{eq:rec} already includes the reconstruction objective in Eq.~\eqref{eq:main} by considering $\|I - \mathrm{Dec}\left(\mathcal{V}_K\right)\|^2=\| \mathbf{x}_1 - \mathrm{Dec}(\mathbf{x}_0, \mathcal{V}_K=\mathcal{V}_{\geq k(0)=1}) \|^2$, because the latter decoder only takes in a new non-informative input: the white noise $\mathbf{x}_0$.
\\
2) \textbf{AR Constraint by Recursive Design}: Due to the correspondence between AR and diffusion recursion in Eq.~\eqref{eq:correspondence}, Eq.~\eqref{eq:rec} is a recursive breakdown of Eq.~\eqref{eq:main} by time-step $t$: $\mathcal{V}_{\geq i}$ is learned from the reconstruction $\| \mathbf{x}_1 - \mathrm{Dec}(\mathbf{x}_t, \mathcal{V}_{\geq k(t)}) \|^2$ that completes the path $\mathbf{x}_t\rightsquigarrow \mathbf{x}_1$; whereas the midway point $\mathbf{x}_t$ encapsulates $\mathcal{V}_{<i}$, which is considered to be already identified by $\mathbf{x}_0\rightsquigarrow \mathbf{x}_t$. This satisfies the probability factorization in Eq.~\eqref{eq:recursive} and the causal structure in Figure~\ref{fig:4a}. 
\\
3) \textbf{AR Constraint by Causal Identification}: To ensure that the learned $\mathcal{V}_K$ is indeed of AR structure, \ie, the encoder \textit{identifies the causal effect} from $\mathcal{V}_{<i}$ to $\mathcal{V}_{\geq i}$, we need to justify that Eq.~\eqref{eq:rec} is an unbiased estimate of $\mathcal{V}_{\geq i}$ from $\mathbf{x}_t$ (\ie, $\mathcal{V}_{<i}$) for all $t\in[0,1]$. To this end, we show that Eq.~\eqref{eq:rec} induces the causal graph in Figure~\ref{fig:4c}: Causation $\mathbf{x}_0 \to \mathbf{x}_t \leftarrow \mathbf{x}_1$ denotes that $\mathbf{x}_t$ is sampled from $q(\mathbf{x}_t|\mathbf{x}_1)$ by mixing noise $\mathbf{x}_0$ and image $\mathbf{x}_1$; causation $\mathbf{x}_t \to \mathcal{V}_{\geq k(t)} \leftarrow \mathbf{x}_1$ denotes that the tokens $\mathcal{V}_{\geq k(t)}$ are learned from $\mathbf{x}_1$ and $\mathbf{x}_t$. In this way, $\mathbf{x}_0$ serves as an \textit{instrument variable} (IV)~\cite{pearl2009causality}, independent of the confounder $\mathbf{x}_1$. Recall the re-parametrization: $\mathbf{x}_t = \sigma(t)\cdot \mathbf{x}_0+\mu(t)\cdot \mathbf{x}_1$, where $\sigma(t)$ and $\mu(t)$ can be considered as time-specific constants~\cite{liuflow}. Thus, the inner expectation of Eq.~\eqref{eq:rec} can be rewritten as:
\begin{equation}
    \mathop{\mathbb{E}}_{\mathbf{x}_0\sim \mathcal{N}(0,1)} \left[ \| \mathbf{x}_1 - \mathrm{Dec}\left(\sigma(t)\cdot \mathbf{x}_0+\mu(t) \cdot \mathbf{x}_1, \mathcal{V}_{\geq k(t)}\right) \|^2 \right],
    \label{eq:new_rec}
\end{equation}
which implies that $\mathcal{V}_{\geq k(t)}$ can be directly estimated from the IV $\mathbf{x}_0$, ensuring that $\mathcal{V}_{\geq k(t)}$ learned from $\mathbf{x}_t$ is unbiased, even in the presence of the confounder $\mathbf{x}_1$.

\begin{figure*}
    \centering
    \footnotesize
    \includegraphics[width=\textwidth]{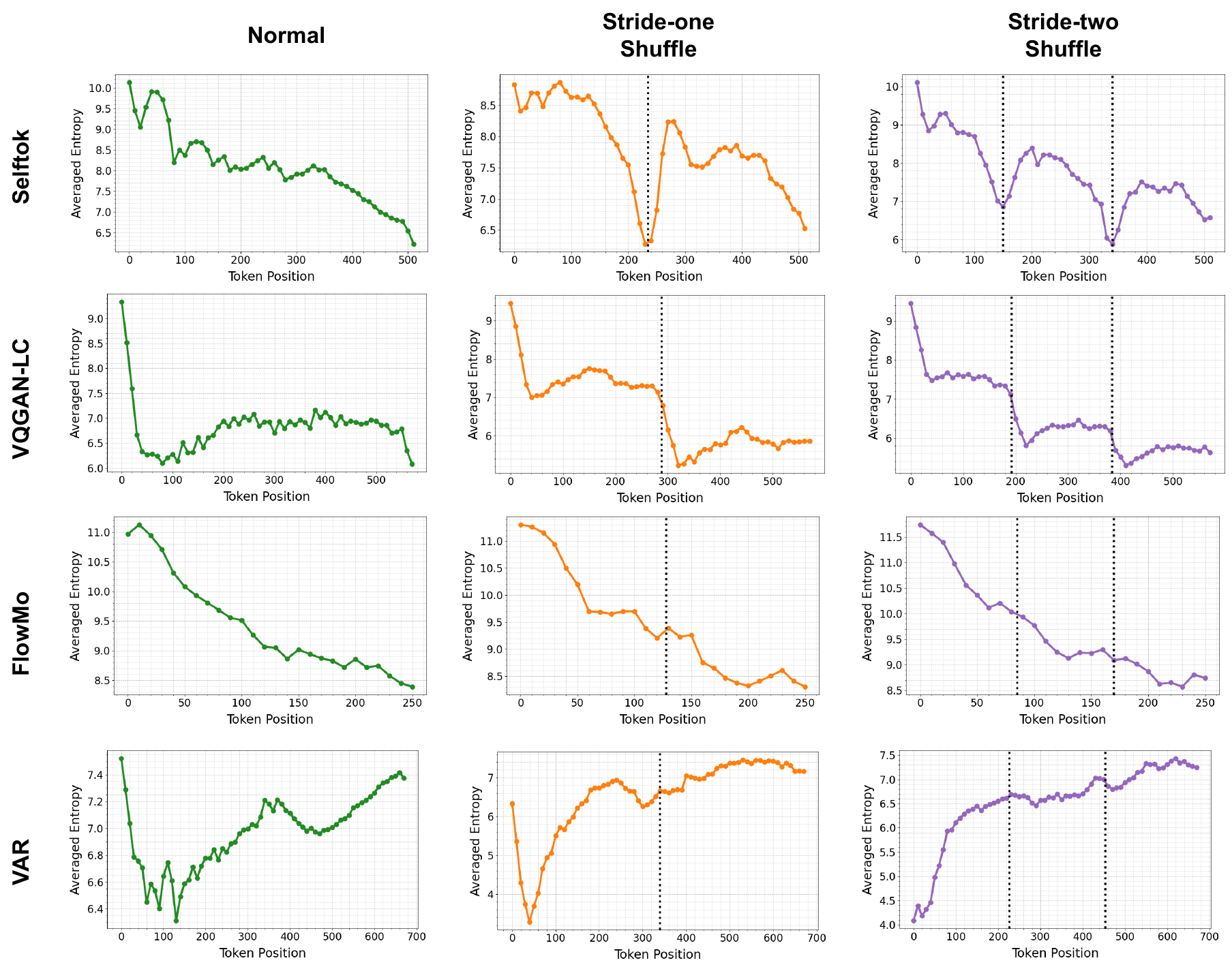}
    \vspace*{-5mm}
    \caption{Plots of the next-token prediction entropy versus token position for our Selftok, 2D spatial tokens (VQGAN-LC~\cite{vqganlc}), 1D tokens (FlowMo~\cite{flowmo}), multi-scale 2D tokens (VAR~\cite{var}), using the original or shuffled sequences. Only Selftok exhibits a segmented decreasing trend that aligns with the three sequence orders. Although VQGAN-LC also displays a segmented trend, each segment is not decreasing. Conversely, while FlowMo shows a decreasing trend, it is not segmented under the shuffled orders.}
    \vspace*{-1mm}
    \label{fig:entropy_token}
\end{figure*}

Now, we empirically verify that the structure of Selftok is AR by plotting the token prediction entropy curves \textit{w.r.t.} token positions under three generation orders using a dAR model (Llama 3.1). Besides the normal sequential order $[v_1,v_2,v_3,...]$, we use another two orders: 1) stride-one shuffle, which is a concatenation of subsequence $[v_{1},v_{3},...]$ followed by subsequence $[v_{2},v_4, v_6, ...] $, and 2) stride-two shuffle, which is a concatenation of subsequence $[v_1,v_4,v_7,...]$, $[v_2,v_5,v_8,...]$, and $[v_3, v_6, v_9, ...]$. The design principle of these orders is simple: an ordered subsequence of an AR sequence is still AR. As entropy measures the uncertainty in token prediction, if the sequence is AR, the entropy trend is generally decreasing. Therefore, if the token sequence is AR, the two shuffled orders should demonstrate a segmented decreasing curve. As shown in Figure~\ref{fig:entropy_token}, we can see that only Selftok demonstrates such a
segmented decreasing trend corresponding to the three sequence orders.

\subsection{Implementation}
\label{sec:2.2}

\begin{figure*}
    \centering
    \footnotesize
    \includegraphics[width=\textwidth]{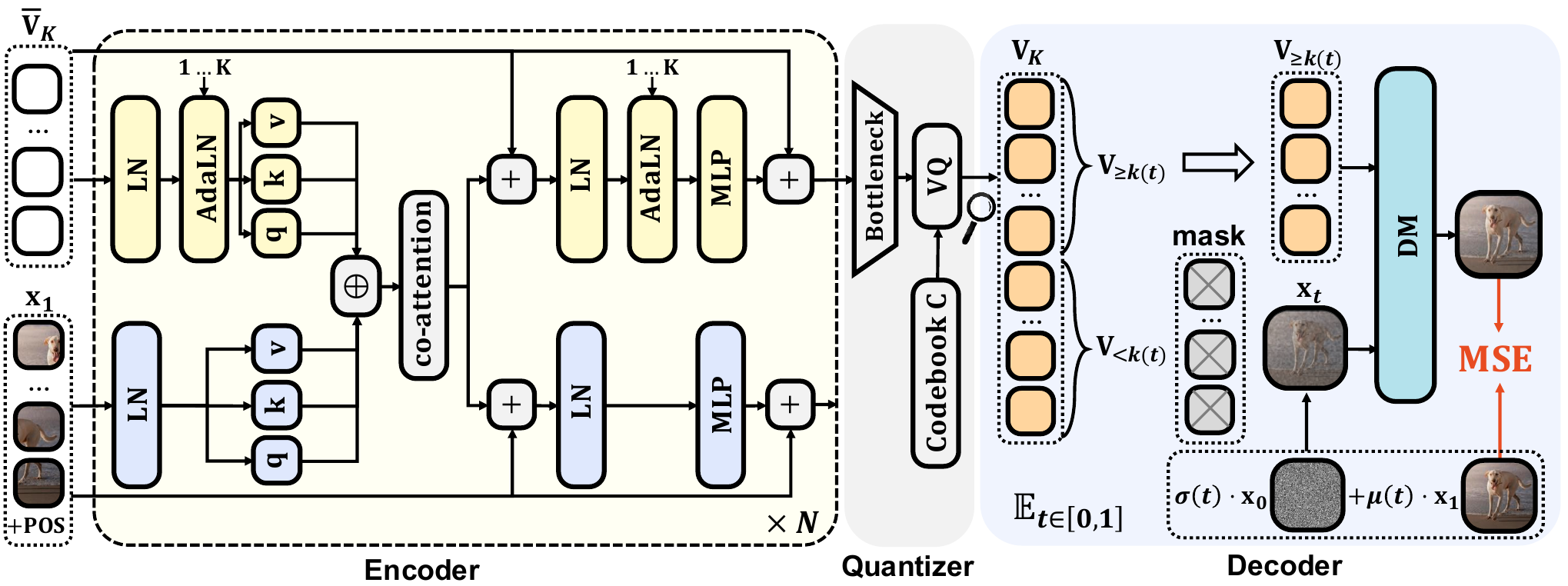}
    \vspace*{-5mm}
    \caption{Selftok architecture diagram. $+\mathbf{POS}$: adding positional embeddings; $\mathbf{LN}$: layernorm; $\mathbf{AdaLN}$: token-aware adaptive layernorm, which differentiates token embeddings; $\mathbf{MSE}$: the mean squared error for the reconstruction objective; $\mathbf{DM}$: pre-trained diffusion model.}
    \vspace*{-1mm}
    \label{fig:tokenizer}
\end{figure*}

Building on the design principles of Selftok outlined above, we now describe the implementation details illustrated in Figure~\ref{fig:tokenizer}\&\ref{fig:renderer}: 1) an encoder that outputs $K$ continuous token embeddings from an input image $I$, 2) a quantizer that looks up the $K$ token indices $\mathcal{V}_K$ from a codebook $\mathbf{C}$, 3) a decoder for reconstructing $I$ given $\mathcal{V}_{\geq k(t)}$ and $\mathbf{x}_t$, and 4) a one-step renderer that leverages the learned Selftok tokens $\mathcal{V}_K$ for fast, one-step reconstruction. 

\subsubsection{Encoder}
\label{sec:2.2.1}

We use a dual-stream transformer backbone like MMDiT~\cite{esser2024scaling}, which consists of an image stream (blue modules) and a token stream (yellow modules). Each stream has its own parameters, specialized for processing patch-based image embeddings and AR-based token embeddings. The backbone consists of $N$ blocks with identical architecture. We elaborate on our design choices below:

\noindent\textbf{Input image}. We use the pre-trained VAE of SD3~\cite{esser2024scaling} to transform the images into VAE latent embeddings. This downscales an image (in pixel-wise RGB values) by a factor of 8, \eg, turning an image of size $256\times256$ to a $32\times 32$ latent.
We follow ViT~\cite{vit} to patchify and flatten the latent into an $\mathbb{R}^{M\times D}$ sequence denoted as $\mathbf{x}_1$, where $M$ is $\#\textrm{patches}$ and $D=512$ is the embedding dimension. We add 2D sinusoidal positional embedding to the sequence before sending it to the image stream.

\noindent\textbf{Input token}. The token stream inputs a sequence of learnable continuous token embeddings $\overline{\mathbf{V}}_K \in \mathbb{R}^{K\times D}$, which is part of the model parameters and independent from $I$.

\noindent\textbf{Co-attention}. In each encoder block, the two streams interact through a co-attention, where their queries, keys, and values are concatenated to compute scaled dot-product attention. As the output of the image stream will not be input to the subsequent quantizer, to improve computational efficiency, we apply an attention mask that prevents queries from the image stream from attending to keys in the token stream. This reduces the attention computation, while still giving the token stream access to image features to complete the encoding.

\noindent\textbf{Token-aware adaptive layernorm (AdaLN)}. Motivated by how MMDiT uses AdaLN to differentiate diffusion time-steps, our AdaLN uses token-aware scale and shift parameters $\boldsymbol{\alpha}_k ,\boldsymbol{\beta}_k \in \mathbb{R}^{D}$ for each token index $k\in\{1,\ldots,K\}$.
Given an input token embedding $\mathbf{v}_k\in\mathbb{R}^{D}$, the AdaLN output is computed as $\boldsymbol{\alpha}_k \odot \mathrm{LN}(\mathbf{v}_k) + \boldsymbol{\beta}_k$, where $\mathrm{LN}$ denotes layer normalization.

\noindent\textbf{Output}. After $N$ blocks, the input learnable token embeddings $\overline{\mathbf{V}}_K$ are eventually transformed into the $K$ continuous embeddings $\mathbf{V} = [\mathbf{v}_1,...,\mathbf{v}_K]\in\mathbb{R}^{K\times D}$ as output. They are subsequently sent to the quantizer to obtain the $K$ discrete tokens $\mathcal{V}_K$. The output of the image stream is discarded.

\subsubsection{Quantizer}
\label{subsec:quantizer}
The quantizer module consists of a bottleneck linear layer, followed by a vector quantization step using a learnable codebook.
The bottleneck maps token embeddings $\mathbf{V}$ from $\mathbb{R}^{K\times D} \to \mathbb{R}^{K\times D'}$, where $D'=16<D$. The codebook has $C=2^{15}$ words and denoted as $\mathbf{C}\in \mathbb{R}^{C \times D'}$.
As illustrated in Figure~\ref{fig:tokenizer}, the bottleneck first compresses the token embeddings $\mathbf{V}$ from  $\mathbb{R}^{K\times D}$ to $\mathbb{R}^{K\times D'}$. Then, each $\mathbf{v}_k\in\mathbf{V}$ is quantized by assigning it to a codebook word $\mathbf{c} \in \mathbf{C}$ with the highest cosine similarity. Thus, the discrete token $v_k$ is the integer index of $\mathbf{c}$.
The quantized embedding output is computed with the straight-through estimator~\cite{vqvae} as $\mathbf{v}_k \leftarrow \mathbf{v}_k + (\mathbf{c} - \mathbf{v}_k)\text{.detach}$, where ``detach'' cuts the gradient flow to $(\mathbf{c} - \mathbf{v}_k)$ in backpropagation. This approximation ensures that the forward pass uses the  quantized embedding $\mathbf{c}$, while the backward pass allows gradients to flow through $\mathbf{v}_k$ to the encoder.

\noindent\textbf{Quantization loss}. The quantization loss $\mathcal{L}_Q$ for learning the codebook $\mathbf{C}$ is:
\begin{equation}
    \mathcal{L}_Q = \underbrace{\sum_{k=1}^K \| \mathbf{v}_k - \mathbf{c}_k \|^2}_{\text{commitment loss}} + \underbrace{\sum_{\mathbf{c}\in\mathbf{C}} \bar{p}_\mathbf{c} \mathrm{log}\; \bar{p}_\mathbf{c},}_{\text{entropy loss}} \text{where } \underbrace{\bar{p}_\mathbf{c} = \mathop{\mathbb{E}}_{\mathbf{V}\sim \mathcal{B} } \mathop{\mathbb{E}}_{\mathbf{v}_k\sim \mathbf{V} } \left[ \frac{ \mathrm{exp}( \mathrm{cos}(\mathbf{v}_k, \mathbf{c}) / \tau ) }{ \sum_{\mathbf{c}' \in \mathbf{C}} \mathrm{exp}( \mathrm{cos}(\mathbf{v}_k, \mathbf{c}') / \tau ) } \right]}_{\text{average assignment probability of word $\mathbf{c}$ in batch $\mathcal{B}$}},
    \label{eq:codebook}
\end{equation}
where the temperature $\tau=0.1$. 
The commitment loss encourages each token embedding $\mathbf{v}_k$ to be close to its assigned word $\mathbf{c}_k$.
The entropy loss encourages diverse codebook usage by promoting uniform word assignment, \ie, the average assignment probability $\bar{p}_\mathbf{c}$ of each word should be close to $1/C$.
We compute $\bar{p}_\mathbf{c}$ in a batch $\mathcal{B}$ by averaging a soft assignment probability based on the cosine similarity $\mathrm{cos}(\mathbf{v}_k, \mathbf{c})$ for each $\mathbf{v}_k \in \mathcal{B}$.

\noindent\textbf{EMA update}. Due to the non-differentiable nature of the quantizer, $\mathbf{C}$ is updated using an exponential-moving-average (EMA) to improve the training stability. For each sample batch, we compute a batch-specific centroid for each codebook word $\mathbf{c}\in\mathbf{C}$, defined as the mean embedding of all \(\mathbf{v}_k\) that are assigned to the word. Let \(\overline{\mathbf{C}} \in \mathbb{R}^{C \times D'}\) denote the matrix of these centroids. The codebook update is: $\mathbf{C}\leftarrow \gamma \mathbf{C} + (1-\gamma) \overline{\mathbf{C}}$, where $\gamma=0.8$.

\noindent\textbf{Dead-code reactivation}.
We observe that some codebook words may fall far from the token embedding distribution during early training and become underutilized.
As a result, they rarely match with any $\mathbf{v}_k$. We identify such words by tracking the EMA of $\bar{p}_\mathbf{c}$ across batches, denoted as $\hat{p}_\mathbf{c}\leftarrow \eta \hat{p}_\mathbf{c} + (1-\eta)\bar{p}_\mathbf{c}$, where $\eta=0.99$. We mark $\mathbf{c}$ as dead code if its EMA $\hat{p}_\mathbf{c}$ falls below a threshold (we use $0.0125/C$). When this occurs, we reactivate $\mathbf{c}$ by assigning it to a randomly selected $\mathbf{v}_k$ from the current batch, pulling it back toward the active embedding space.

\subsubsection{Decoder}
\label{sec:2.2.3}

\noindent\textbf{Decoder architecture}.
Our $\mathrm{Dec}$ is a diffusion model initialized from SD3~\cite{esser2024scaling}.
It is a dual-stream transformer MMDiT architecture with the following customization:
1) Our token stream replaces the input of the original language token stream with the quantized embeddings $[\mathbf{v}_{k(t)},\ldots,\mathbf{v}_K]\in \mathbb{R}^{(K-k(t)+1)\times D'}$ (the embeddings of $\mathcal{V}_{\geq k(t)}$) with  1D sinusoidal positional embedding;
2) To remove the original language influence and better adapt to Selftok tokens, the weights of our token stream are trained from scratch;
3) For the AdaLN layers in the image stream, we remove their dependency on the pooled CLIP text embedding and condition them only on the timestep $t$. For the token stream, we continue to use the token-aware AdaLN as in the encoder.

\noindent\textbf{Overall objective}. We jointly optimize the parameters of $\mathrm{Enc}$ (the combination of the encoder  and the quantizer) and $\mathrm{Dec}$, using the objective below:
\begin{equation}
\mathop{\mathrm{min}}_{\substack{\mathrm{Enc}(\mathbf{x}_1) = \mathcal{V}_K, \\ \mathrm{Dec}}} ~~ \mathop{\mathbb{E}}_{t\in [0,1]} \;\mathop{\mathbb{E}}_{\mathbf{x}_0\sim \mathcal{N}(0,1)} \left\|\mathbf{x}_1 - \mathrm{Dec}\left((1-t)\cdot \mathbf{x}_0+t\cdot\mathbf{x}_1, \mathbf{V}_{\geq k(t)}\right) \right\|^2 + \mathcal{L}_Q,
    \label{eq:selftok_overall}
\end{equation}
where $\mathbf{V}_{\geq k(t)}$ are the token embeddings of $\mathcal{V}_{\geq k(t)}$. We highlight the following details:
\\
\textbf{1)} We uniformly sample $t\in[0,1]$, which yields the best performing model. We include ablation results in Section~\ref{sec:2.3.2}.
\\
\textbf{2)} We follow SD3 to use the re-parameterization $\mathbf{x}_t = (1-t)\cdot \mathbf{x}_0 + t\cdot \mathbf{x}_1$, \ie, $\sigma (t) = 1-t$ and $\mu(t) = t$ in Eq.~\eqref{eq:new_rec}.
\\
\textbf{3)} Before sending $\mathbf{V}_{\geq k(t)}$ to the decoder, we implement a re-weighting mechanism by updating each $\mathbf{v}_k$ as below:
\begin{equation}
    \mathbf{v}_k \leftarrow \frac{\alpha}{p_k} \mathbf{v}_k + (\mathbf{v}_k - \frac{\alpha}{p_k}\mathbf{v}_k).\mathrm{detach}, \; \text{where } p_k = \mathbb{E}_{t\in [0,1]} \mathbb{I}(v_k \in \mathcal{V}_{\geq k(t)}),
    \label{eq:reweight}
\end{equation}
where $\alpha=0.5$ is is a weight constant, $p_k$ computes the probability of $v_k \in \mathcal{V}_{k(t)}$ for a uniformly sampled $t\in[0,1]$, and $\mathbb{I}(\cdot)$ is an indicator function.
This re-weighting is introduced to correct the imbalance of $p_k$: as $k$ decreases, ${v}_k$ is more likely to be masked out by ``$\geq$'' in $\mathcal{V}_{\geq k(t)}$, resulting in smaller $p_k$ and fewer gradient updates on $\mathbf{v}_k$.
By applying Eq.~\eqref{eq:reweight}, the value of each embedding $\mathbf{v}_k$ remains unchanged in the forward pass, but in the backward pass, its gradient is scaled inversely to $p_k$ to compensate for the imbalanced $p_k$, \ie, $\mathbf{v}_k$ with fewer gradient updates is assigned with a larger gradient scale.
\\
4) We discuss the choice of token schedule $k(t)$ below and include its ablation in Section~\ref{sec:2.3.2}.

\subsubsection{Token Schedule}
\label{sec:2.2.4}

Recall that the AR constraint in Eq.~\eqref{eq:main} requires that  every token must conform to the decomposition $P(\mathcal{V}_K) \overset{\text{AR}}{=} P(\mathcal{V}_{<i}) \cdot P(\mathcal{V}_{\geq i}|\mathcal{V}_{<i}), \forall i\in[1,K+1]$.
We achieve this decomposition by diffusion time-steps, thanks to the recursive nature of the reverse diffusion process in Eq.~\eqref{eq:correspondence}, denoted as $\mathcal{V}_{\geq i}\Leftrightarrow \mathbf{x}_t\rightsquigarrow \mathbf{x}_1$ and $\mathcal{V}_{< i}\Leftrightarrow \mathbf{x}_0\rightsquigarrow \mathbf{x}_t$. That is to say, the second-half tokens $\mathcal{V}_{\geq i}$ can be learned recursively by the diffusion decoder, conditioned on $\mathbf{x}_t$, which represents the already identified first-half tokens $\mathcal{V}_{< i}$. 
As we uniformly sample $t\in[0,1]$ in training,  the best token schedule should be a uniform assignment $k^*(t)=\ceil{t\times K}+1$ to ensure that every token is involved in the recursive diffusion time-step.
To better understand this, we provide three failure cases:
\\
\textbf{1)} If we allocate all the tokens to $\mathcal{V}_{\geq 1}$, \ie, $k(t) = 1$, $\forall t\in [0,1)$, this corresponds to a trivial decomposition $P(\mathcal{V}_{<1}=[])\cdot P(\mathcal{V}_K|\mathcal{V}_{<1}=[])$, $\mathcal{V}_{K}\Leftrightarrow \mathbf{x}_0\rightsquigarrow \mathbf{x}_1$, and $[]\Leftrightarrow \mathbf{x}_0$, where we always input the full $\mathcal{V}_K$ to the decoder.
So, $\mathcal{V}_K$ loses all the AR property. This case reduces to the FlowMo approach~\cite{flowmo}.
\\
\textbf{2)} If we always allocate all tokens to $\mathcal{V}_{< 1}$, \ie, $k(t)=K+1,\forall t\in (0,1]$, this corresponds to another trivial decomposition $P(\mathcal{V}_K)\cdot P([]|\mathcal{V}_K)$, $[]\Leftrightarrow \mathbf{x}_0\rightsquigarrow \mathbf{x}_1$, and $\mathcal{V}_{K}\Leftrightarrow \mathbf{x}_0$, where we always send an empty sequence to the decoder. This case reduces to the unconditional diffusion generation without learning $\mathcal{V}_K$ at all.
\\
\textbf{3)} Consider a non-extreme case where $k(t)$ is not uniformly aligned with $t$, \eg, $k(t=0.8)=\ceil{0.2\times K}$, we disrespect the decomposition because the majority of tokens $\mathcal{V}_{\geq \ceil{0.2\times K}}$ corresponds to dense time-steps in the short interval $t\in[0.8,1]$, while the rest ones in $\mathcal{V}_{< \ceil{0.2\times K}}$ corresponds to sparse time-steps, violating the balanced recursive correspondence in Eq.~\eqref{eq:correspondence}. 

However, in practice, we empirically observe a better reconstruction quality by designing a schedule $k(t)$ that allocates fewer tokens to smaller $t$, \ie, $k(t) < k^*(t)$ for $t<0.5$.
This aligns with the well-known trait of diffusion models: the early path $\mathbf{x}_0\rightsquigarrow \mathbf{x}_t$ for a small $t$ has minimal impact on the reconstruction $\mathbf{x}_t \rightsquigarrow \mathbf{x}_1$, which can be omitted~\cite{wang2024closer,panvr}.
Nevertheless, identifying the optimal $k(t)$ that effectively balances the AR property and diffusion generation quality remains a challenging open problem, which we leave for future work.

\subsubsection{One-step Renderer}
\label{sec:2.2.5}

As observed from the training objective in Eq.~\eqref{eq:selftok_overall},
Selftok tokens $\mathcal{V}_K$ participate in the reconstruction at every diffusion time-step.
Hence after training, we can apply a standard multi-step diffusion sampler~\cite{esser2024scaling,flowmo} to decode $\mathcal{V}_K$ into a reconstructed image. However, this process is slow as it requires multiple sequential forward passes.

To accelerate this, we leverage the fact that $\mathcal{V}_K$ discretizes the entire generative path $\mathbf{x}_0 \rightsquigarrow \mathbf{x}_1=I$, \ie, it encodes all the information of $I$, up to quantization loss.
Hence, we aim to build a renderer $R(\mathcal{V}_K)$ that reconstructs $I$ in a single forward pass.
We initialize $R$ with the decoder weights optimized by Eq.~\eqref{eq:selftok_overall}. To remove its dependency on $\mathbf{x}_t$, we replace it with a sequence of learnable ``canvas'' token embeddings as shown in Figure~\ref{fig:renderer} (b), which becomes part of the model parameters of $R$.
Then with the learned token embeddings $\mathbf{V}_K = \mathrm{Enc}(I)$ frozen, we optimize $R$ jointly with an MSE loss for pixel-level reconstruction,  LPIPS~\cite{lpips} and GAN~\cite{gan} loss for perceptual quality, as including the latter two resolves the well-known blurry reconstruction issue when training a decoder with the MSE loss alone~\cite{stylegan, vqgan}:
\begin{equation}
\mathop{\mathrm{min}}_{R(\mathbf{V}_K) = I'} \, \mathop{\mathrm{max}}_{D} \left[ \underbrace{\| I - I' \|^2}_{\text{MSE loss}} + \underbrace{ \lambda_1 \mathrm{LPIPS}(I,I')}_{\text{perceptual loss}} + \underbrace{\lambda_2 \left(\log D(I) + \log (1-D(I')\right)}_{\text{GAN loss}} \right],
    \label{eq:renderer}
\end{equation}
where $\lambda_1,\lambda_2$ are loss weights, $D$ is the discriminator of the GAN.
To improve training stability, we set $\lambda_1=0.1,\lambda_2=0$ for the first 30k training iterations and $\lambda_1=0.5,\lambda_2=0.5$ afterwards. As shown in Figure~\ref{fig:renderer} (a), besides the improved visual perception, the one-step renderer brings two benefits: 1) it significantly reduces the image generation time, and 2) it eliminates the randomness introduced by the random seed in diffusion-based generation (see Figure~\ref{fig:renderer} (a)). Explicit experimental results are shown in Section~\ref{sec:2.3.2}.

\subsection{Validation}

\label{sec:2.3}

In this section, we empirically demonstrate that the Selftok tokenizer achieves the best reconstruction quality over all baselines in Section~\ref{sec:2.3.1}, while encoding tokens that satisfy the AR property (Figure~\ref{fig:entropy_token}).
In addition, we provide comprehensive ablation studies in Section~\ref{sec:2.3.2} to analyze various design choices, including \#tokens $K$, codebook size $C$, token schedule $k(t)$, time-step sampler and the use of one-step renderer.

\noindent\textbf{Settings}.
We used the train split of ImageNet-1k dataset~\cite{deng2009imagenet} to train our tokenizer, where images were resized and center-cropped to the size of 256$\times$256. For evaluation, we used the validation split with the same image pre-processing.
We evaluate the reconstruction quality using four metrics: 1) rFID~\cite{rfid} measures the dissimilarity between the distribution of reconstructed validation split and ground-truth train split; 2) PSNR measures a pixel-wise MSE between each image $I$ and its reconstruction $I'$; 3) SSIM measures the perceptual similarity between $I$ and $I'$ using a statistical approach~\cite{ssim}; 4) LPIPS does so by leveraging a pre-trained network~\cite{lpips}.

\subsubsection{Main Results}
\label{sec:2.3.1}

\begin{figure*}
    \centering
    \footnotesize
    \includegraphics[width=\textwidth]{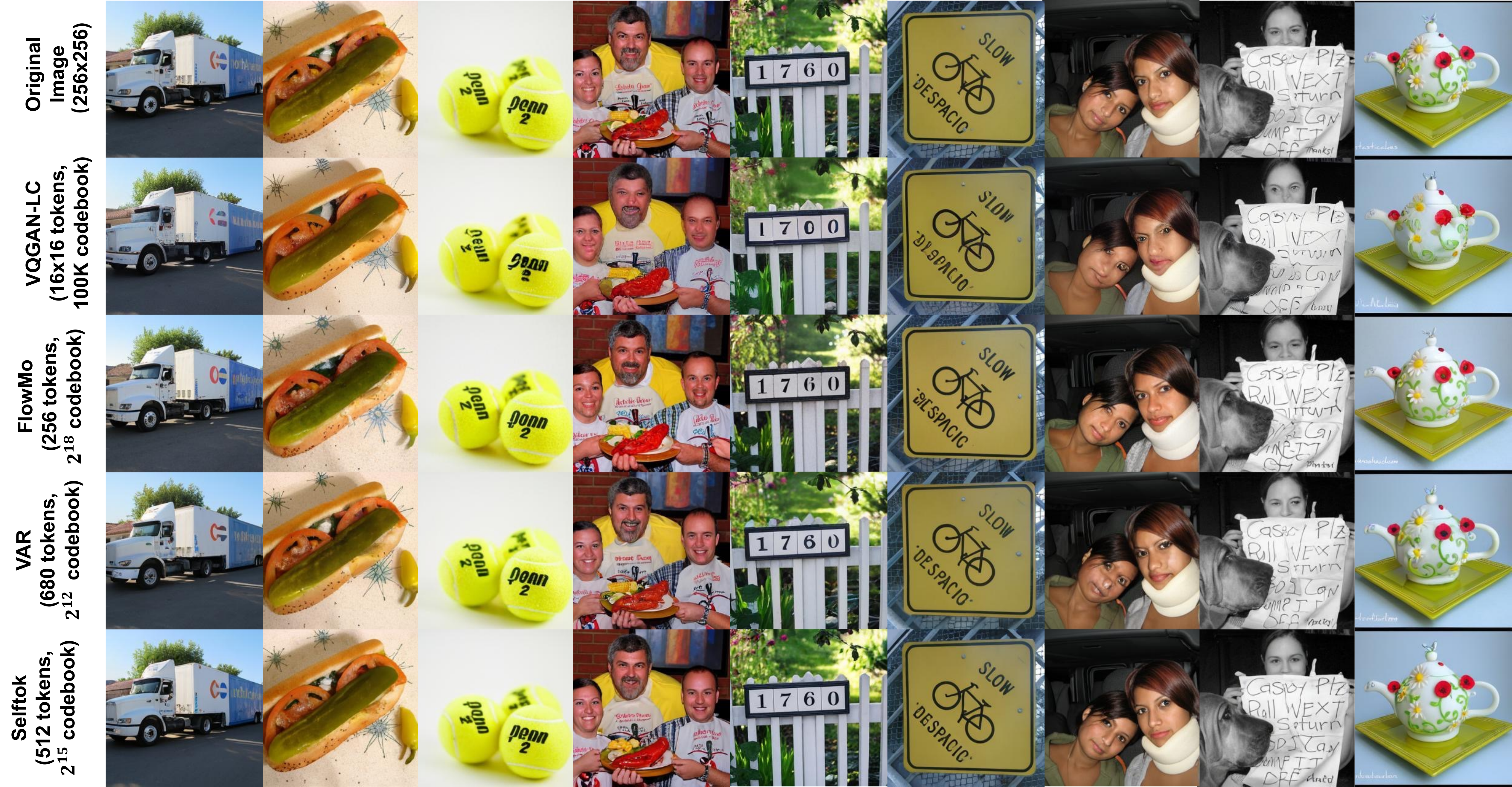}
    \vspace*{-5mm}
    \caption{Comparison of reconstructions from different tokenizers.}
    \vspace*{-1mm}
    \label{fig:reconstruction}
\end{figure*}

The results, presented in Table~\ref{tab:tokenizer}, show that Selftok achieves SoTA reconstruction performance (in terms of rFID, PSNR, SSIM, and LPIPS) and outperforms existing spatial tokenizers (both 1D and 2D, such as TiTok~\cite{titok} and VAR~\cite{vqganlc}) as well as tokenizers using diffusion models as decoders, such as FlowMo~\cite{flowmo}. Specifically, we can observe that when using 1024 tokens, existing tokenizers like Cosmos and FlowMo have worse reconstruction performance compared to ours (with a PSNR difference greater than 1). We also show the qualitative comparison between different tokenizers in Figure~\ref{fig:reconstruction}.
Notably, we verify in Figure~\ref{fig:disentangle} that Selftok tokens are non-spatial by visualizing 1) progressive reconstruction by providing the decoder with an expanding sequence of tokens $\mathcal{V}_{\geq k(t)}$ for a decreasing $t$ from left to right, and 2) interpolation by gradually changing $\mathcal{V}_{\geq k(t)}$ for a decreasing $t$ of the left image with that of the right image.
We further demonstrate the AR property of Selftok tokens in Figure~\ref{fig:entropy_token}.
\begin{table*}
    \centering
    \small
    \begin{tabular}{llrrrrrr}
    \toprule
    \textbf{Tokenizer} & 
    \textbf{Type} &
    \textbf{\#Token} & 
    \textbf{\#Code} & 
    \textbf{rFID$\downarrow$} & 
    \textbf{PSNR$\uparrow$} & 
    \textbf{SSIM$\uparrow$} & 
    \textbf{LPIPS$\downarrow$} \\ \midrule
    LlamaGen \cite{llamagen} & 2D & 16$\times$16 & $2^{14}$ &  2.19 & 20.67 & 0.589 & 0.132 \\
    MaskBiT$^{\dagger}$\cite{maskbit} & 2D &16$\times$16 & $2^{14}$ &  1.37 & 21.50 & 0.560 & - \\ 
    Cosmos \cite{agarwal2025cosmos}& 2D & 16$\times$16 & $\approx 2^{16}$ &  4.40 & 19.98 & 0.536 & 0.153 \\
    VQGAN-LC$^{\dagger}$~\cite{vqganlc}  & 2D & 16$\times$16 & $100,000$ &  2.62 & 23.80 & 0.589 & 0.120 \\ 
    OpenMagViT-V2 \cite{openmagvit_v2} & 2D & 16$\times$16 & $2^{18}$ &  1.17 & 21.63 & 0.640 & 0.111 \\ 
    ViT-VQGAN$^{\dagger}$ \cite{vitvqgan} & 2D &32$\times$32 & $2^{13}$ &  1.28 & - & - & - \\
    LlamaGen \cite{llamagen} & 2D &32$\times$32 & $2^{14}$ &  0.59 & 24.44 & 0.768 & 0.064 \\
    Cosmos \cite{agarwal2025cosmos} & 2D &32$\times$32 & $\approx 2^{16}$ &  0.87 & 24.82 & 0.763 & 0.070 \\
     VAR~\cite{var} &2D & 680 & $2^{12}$  & 0.99 &  22.12 & 0.624 & 0.109 \\ 
    \midrule
     TiTok-L-32 \cite{titok} & 1D & 32 & $2^{12}$ & 2.21 & 15.60 & 0.359 & 0.322 \\ 
    TiTok-B-64 \cite{titok} & 1D &64 & $2^{12}$ &  1.70 & 16.80 & 0.407 & 0.252 \\
    TiTok-S-128 \cite{titok} & 1D &128 & $2^{12}$ &  1.71 & 17.52 & 0.437 & 0.210 \\
    FlexTok~\cite{flextok} &1D & 256 & $64,000$ & 1.45 & 18.53 & 0.465 & 0.222 \\
    FlowMo-Lo$^{\dagger}$~\cite{flowmo} &1D & 256 & $2^{18}$ &  0.95 & 22.07 & 0.649 & 0.113 \\
    FlowMo-Hi$^{\dagger}$~\cite{flowmo} & 1D &1,024 & $2^{14}$ &  0.56 & 24.93 & 0.785 & 0.073 \\
    \midrule
    Selftok (Ours) &1D & 512 & $2^{15}$ &  \textbf{0.70} & \textbf{24.14} & \textbf{0.709} & \textbf{0.084} \\ 
    Selftok (Ours) &1D & 1,024 & $2^{15}$ &  \textbf{0.54} & \textbf{26.30} & \textbf{0.805} & \textbf{0.063} 
    \\ \bottomrule
    \end{tabular}
    \vspace{-2mm}
    \caption{Reconstruction performance of different tokenizers on $256\times 256$-resolution ImageNet 50k validation set.
    $^{\dagger}$\hspace{-0.3mm} Results from the original paper.
    }
    \label{tab:tokenizer}
\end{table*}

\subsubsection{Ablations}
\label{sec:2.3.2}

\noindent\textbf{\#Tokens $K$ and codebook size $C$.}
Increasing $K$ or $C$ trades off compression rate in favor of reconstruction quality. We verify this in Table~\ref{tab:renderer} and Table~\ref{tab:codebook}, where we tried $K$ of 512 and 1,024, and $C$ of 32,768 ($2^{15}$) and 65,536 ($2^{16}$). For a lower compression rate, we chose $C$=32,768.
\begin{figure*}[t]
    \centering
    \begin{minipage}{0.63\linewidth}
        \centering
        \small
         \begin{table}[H]
            \centering
            \resizebox{\textwidth}{!}{
            \begin{tabular}{llrrr}
                \toprule
                \textbf{Time sampl.} & 
                \textbf{Token sched.} & 
                \textbf{PSNR$\uparrow$} & 
                \textbf{SSIM$\uparrow$} & 
                \textbf{LPIPS$\downarrow$} 
                \\ \midrule
                uniform      & custom   & \textbf{21.86} & \textbf{0.600} & \textbf{0.150}   \\ \midrule
                uniform      & uniform   & 21.10 & 0.564 & 0.177 \\ 
                uniform      & logit-normal  & 20.78 & 0.555 & 0.180 \\
                logit-normal & custom   & 20.98 & 0.561 & 0.170  \\
                logit-normal & uniform      & 19.89 & 0.498 & 0.205 \\
                logit-normal & logit-normal & 20.08 & 0.513 & 0.196 \\
                \bottomrule
            \end{tabular}}
            \caption{Ablation on time sampler and token schedules. `sampl.' and `sched.' denote `sampler' and `schedule'.}
            \label{tab:schedule}
        \end{table}
    \end{minipage}
    \hfill
    \begin{minipage}{0.35\linewidth}
        \centering
        \includegraphics[scale=0.24]{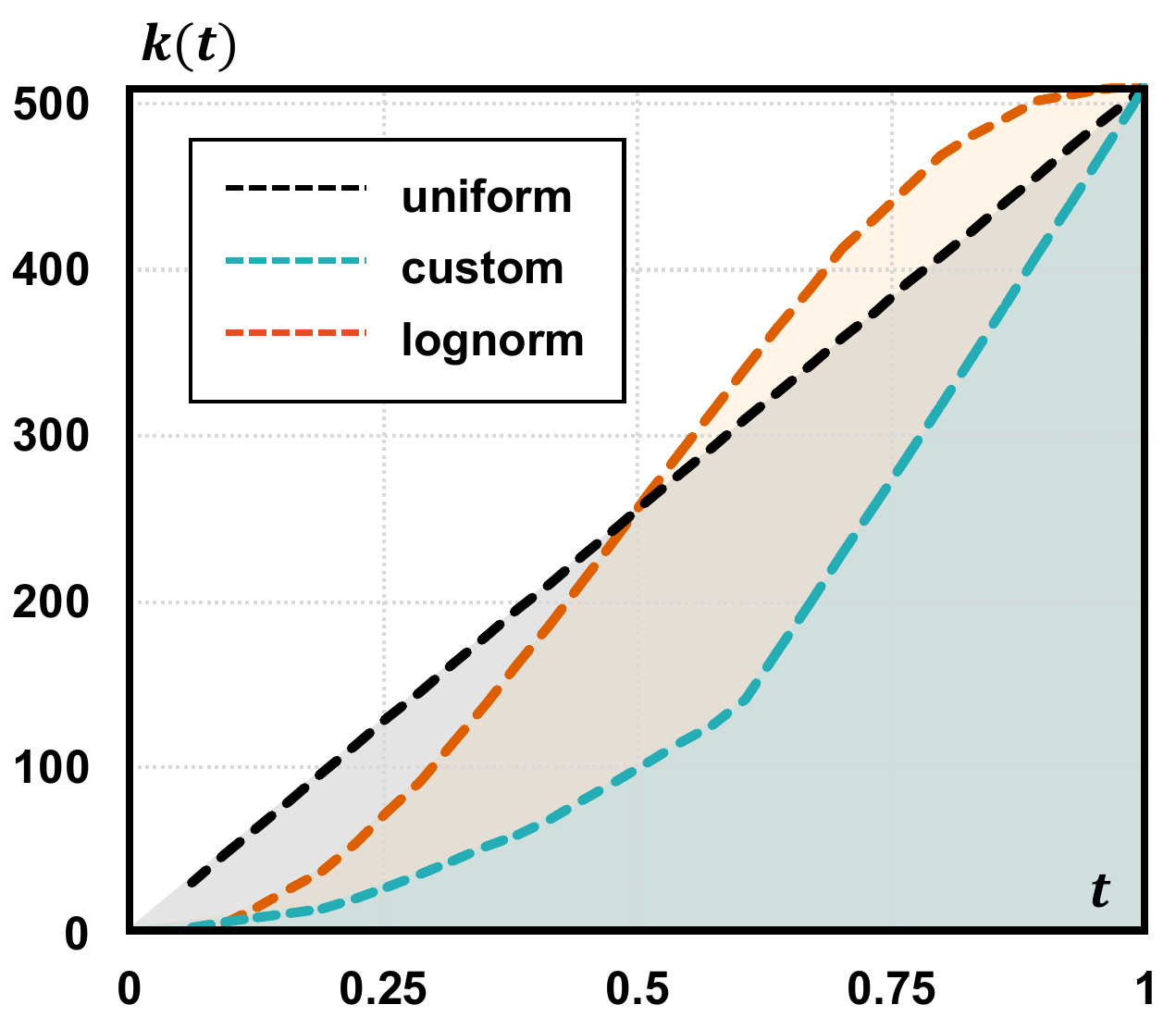}
        \caption{Token schedule $k(t)$. ``lognorm'' denotes logit-normal.}
        \label{fig:tk_curve}
    \end{minipage}
\end{figure*}

\begin{wraptable}{r}{0.42\textwidth}
    \begin{tabular}{lrrrrr}
    \toprule
    \textbf{\#Code} & 
    \textbf{PSNR$\uparrow$} & 
    \textbf{SSIM$\uparrow$} & 
    \textbf{LPIPS$\downarrow$} \\ \midrule
            $2^{15}$     & 21.86 & 0.600 & 0.150 \\
                 $2^{16}$   & 22.12 & 0.618 & 0.135
    \\ \bottomrule
    \end{tabular}
    \vspace{-2mm}
    \caption{Ablation on the codebook size without one-step renderer. 
    }
    \label{tab:codebook}
    \vspace{-1em}
\end{wraptable} 
\noindent\textbf{Time sampler}. For time sampler, besides the simple uniform sampling, SD3~\cite{esser2024scaling} introduces the logit-normal time-step sampler by assigning higher probability density to mid-range time-steps ($t\approx 0.5$). We compared the reconstruction performance when using uniform and logit-normal sampling in Table~\ref{tab:schedule}, which shows that the simple uniform sampling performs the best for Selftok.

\noindent\textbf{Token schedule $k(t)$}. We explored three different choices for \( k(t) \): 1) the uniform one with $k(t)=\ceil{t\times K}+1$; 2) a custom schedule that allocates few tokens to small $t$; and 3) a logit-normal schedule that allocates few tokens to both small and large $t$. We plot $k(t)$ in Figure~\ref{fig:tk_curve} and compare the performance of the models trained with each schedule in Table~\ref{tab:schedule}. As we mentioned earlier in Section~\ref{sec:2.2.4}, our manually designed schedule wins as it aligns with the trait of diffusion models, \ie, $\mathbf{x}_0\rightsquigarrow \mathbf{x}_t$ for a small $t$ has minimal impact on reconstruction, hence tokens should be allocated elsewhere.

\begin{wraptable}{r}{0.60\textwidth}
    \begin{tabular}{clrrr}
    \toprule
    \textbf{Model} & 
    \textbf{Tokens} & 
    \textbf{PSNR$\uparrow$} & 
    \textbf{SSIM$\uparrow$} & 
    \textbf{LPIPS$\downarrow$} \\ \midrule
    \multirow{2}{*}{\shortstack{Selftok\\w/o renderer}} & 512     & 21.86 & 0.600 & 0.150 \\
                                                       & 1,024   & 23.06 & 0.670 & 0.119 \\ \midrule
    \multirow{2}{*}{\shortstack{Selftok\\w/ renderer}}  & 512     & 24.14 & 0.709 & 0.084 \\
                                                       & 1,024   & 26.26 & 0.805 & 0.063 \\
    \bottomrule
    \end{tabular}
    \vspace{-2mm}
    \caption{Ablation on the one-step renderer.}
    \label{tab:renderer}
    \vspace{-1.5em}
\end{wraptable} 

\begin{figure*}
    \centering
    \footnotesize
    \includegraphics[width=\textwidth]{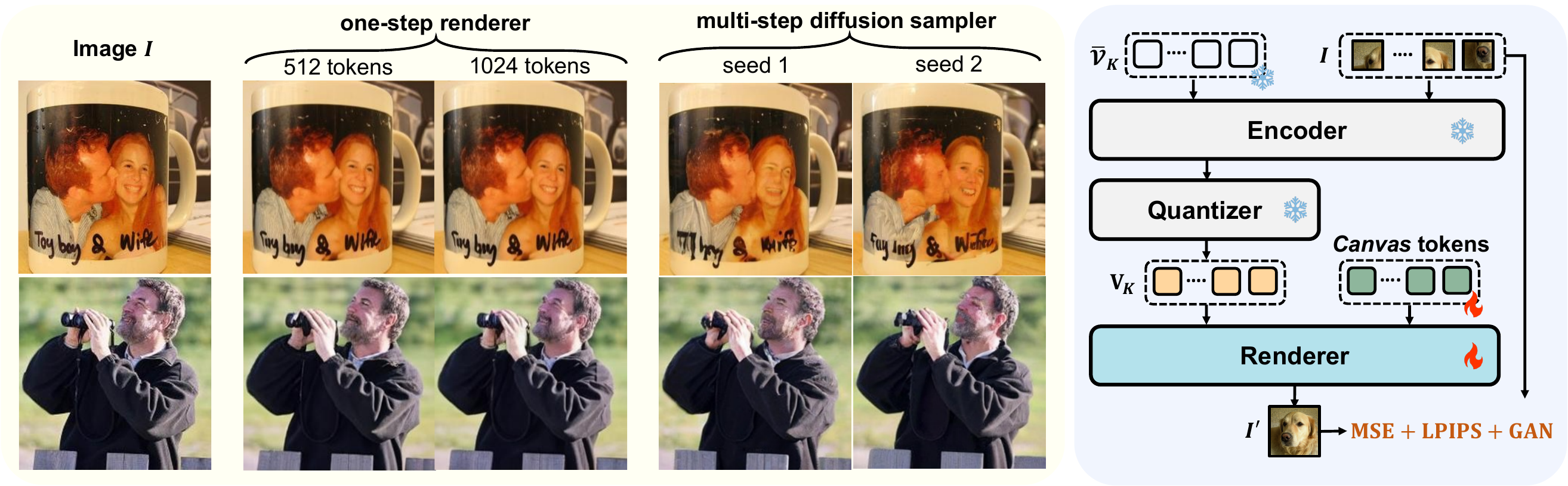}
    \vspace*{-5mm}
    \caption{Left: (a) Reconstructions with one-step renderer (512 or 1024 tokens) and multi-step diffusion sampler (512 tokens, two seeds); Right: (b) Renderer architecture diagram.}
    \vspace*{-1mm}
    \label{fig:renderer}
\end{figure*}

\noindent\textbf{One-step Renderer}. We trained a one-step renderer with Eq.~\eqref{eq:renderer} to further accelerate the image reconstruction process. With the one-step renderer, the reconstruction time decreases from 8.2 sec/per image to \textbf{0.31 sec/per} image (the reconstruction time is tested in one Huawei Asecnd 910B2 with batchsize $=1$). Additionally, the reconstruction performance improves, as shown in Table.~\ref{tab:renderer} and we also present some qualitative results before and after using the one-step renderer in Figure~\ref{fig:renderer} (b). One-step render helps both accelerate the reconstruction speed and improves the reconstruction quality in a large scale. 

\section{Selftok-based VLM}
\label{sec:3}

Thanks to the Selftok tokenizer, images can be encoded into discrete token sequences, which are then processed by a single, purely discrete AR model for visual generation and reasoning. Each visual token sequence can be decoded back into pixel space using the one-step renderer. In Section~\ref{sec:3.1}, we show the model architecture and training strategy of VLM. Then we present validation results in Section~\ref{sec:3.2}.

\subsection{Training}
\label{sec:3.1}
\begin{figure*}[!htb]
    \centering
    \begin{minipage}[b]{\textwidth}
         \includegraphics[width=\textwidth]{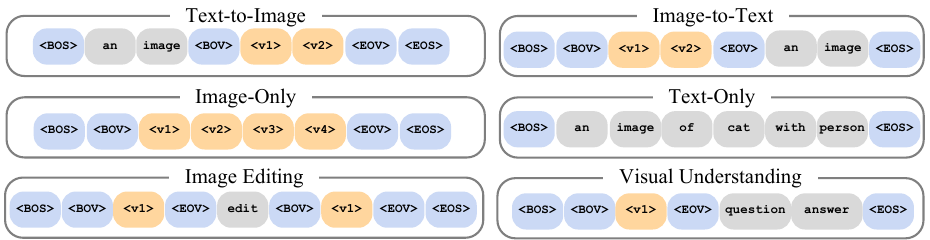}
         \caption{Illustration of the proposed data format for cross-modality and cross-task alignment.}
         \label{fig:ar}
    \end{minipage}
    \begin{minipage}[t]{\textwidth}
          \includegraphics[width=\textwidth]{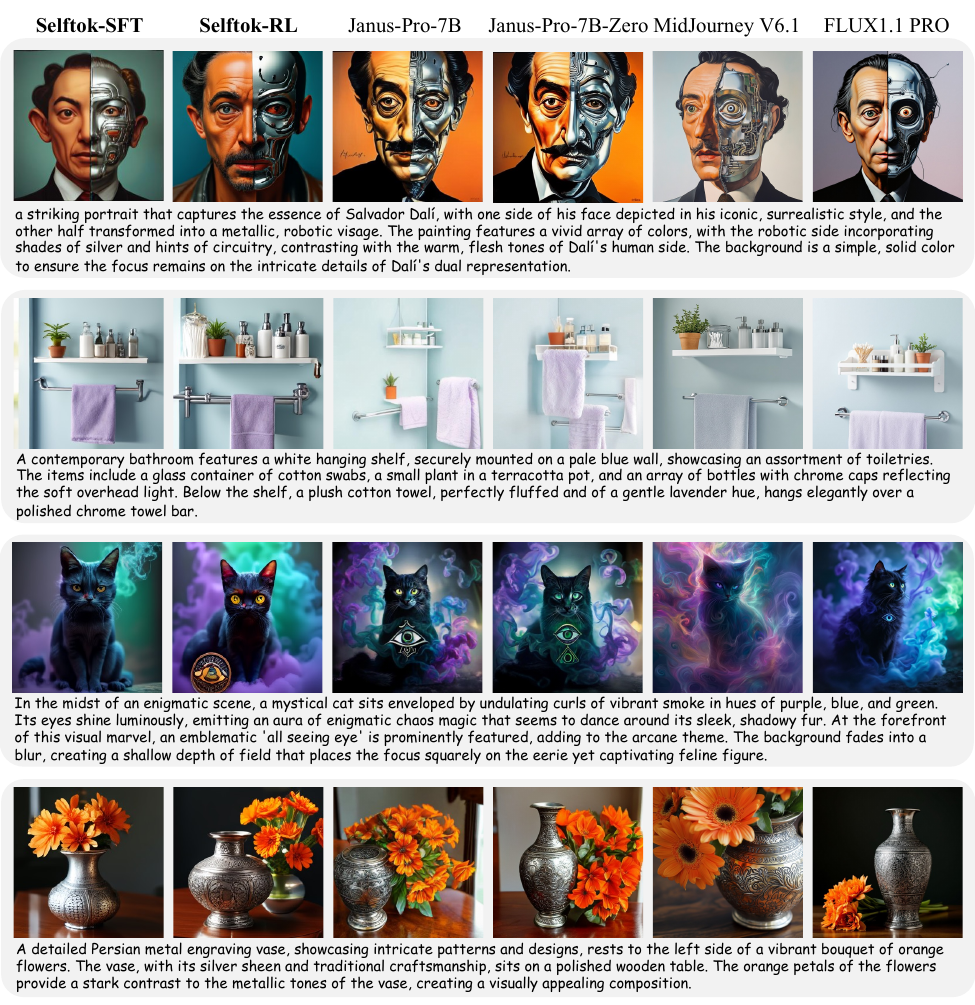}
          \caption{Text-to-Image generation results by Selftok using the text prompts of DPG-Bench.}
          \label{fig:t2i_example}
    \end{minipage}
\end{figure*}

We initialize the VLM from the pretrained Llama3-8B~\cite{llama} model and expand its vocabulary with an additional 32,768 Selftok visual words. As a result, the model's vocabulary integrates both textual and visual tokens into a unified embedding space.
As illustrated in Figure~\ref{fig:1a}, the VLM is trained using the standard language modeling objective, which aims to maximize the log-likelihood of multimodal token sequences in an AR fashion:
\[
P(\mathcal{Y}) = \sum_{i=1}^{|\mathcal{Y}|} \log P_\theta(y_i | \mathcal{Y}_{<i}),
\]
where the sequence \(\mathcal{Y}\) may consist of interleaved language and visual tokens, and thus \(y_i\in\mathcal{Y}\) denotes either a language token \( \langle w_i \rangle \) or a visual token \( \langle v_i \rangle \).  Since both text and image content are represented as discrete token IDs, the prediction head is shared and supervised at each position using a cross-entropy loss. The training consists of the following two stages:

\noindent\textbf{Stage1: Cross-modality Alignment.} 
In this stage, we aim to learn the alignment between visual tokens and language tokens, thereby facilitating the transition of the pre-trained Llama3 model from LLM to VLM. To achieve this, we introduce four data formats designed to address the challenges of cross-modality alignment. Each format helps the model process and integrate vision and language inputs for coherent multimodal understanding and generation. The \textit{Text-to-Image} format aligns caption with visual data, enabling image generation from textual descriptions. Conversely, the \textit{Image-to-Text} format facilitates understanding tasks by associating visual data with textual descriptions. To address potential misalignments that can occur during text-to-image tasks, the \textit{Image-Only} format is introduced, allowing the model to learn visual structure independently. Finally, the \textit{Text-Only} data ensures the preservation of the model's linguistic capabilities, maintaining its ability to process and generate text. These formats and their functions are summarized in Figure~\ref{fig:ar}, with special tokens such as \texttt{[BOS]} and \texttt{[EOS]} marking the sequence boundaries, and \texttt{[BOV]} and \texttt{[EOV]} indicating the start and end of visual data. The training data is comprised of 530 million high-quality image-text pairs and text sequences (See Appendix for more details).

\noindent\textbf{Stage2: Cross-task Alignment.} In this stage, we perform cross-task alignment to enable the model to learn human instructions across various tasks. This is accomplished through supervised fine-tuning (SFT) on datasets from three distinct tasks: 1) text-to-image generation, 2) image editing, and 3) image understanding. The instruction format follows the structure \texttt{``USER: <Instructions> ASSISTANT: <Answers>''}, where only the content of \texttt{<Answer>} contributes to the loss function, optimizing the model's ability to provide accurate responses. Detailed data organization and training details are provided in the Appendix.

\subsection{Validation}
\label{sec:3.2}
After completing the two stages of training, we evaluated the model's performance on three downstream tasks: text-to-image generation, image editing, and image understanding. During inference, we introduced logit adjustment to improve model performance. The adjustment is applied when the entropy of the conditional logits exceeds a threshold $\tau$, indicating that the model is uncertain about the current token: 
\begin{equation}
\text{logit}_{\text{adjusted}} = 
\begin{cases} 
\text{logit}_c + (\gamma - 1) \cdot (\text{logit}_c - \text{logit}_u), & \text{if} \; H(\text{logit}_c) > \tau \\
\text{logit}_c , & \text{otherwise}
\end{cases}
\end{equation}
By mitigating this uncertainty, the adjustment strengthens the model’s performance across diverse tasks. Logit$_c$ and logit$_u$ represent the logits predicted by the model for each token in the autoregressive sequence generation process. For \textit{text-to-image generation}, $\text{logit}_c$ refers to the output with text prompt input, while $\text{logit}_u$  refers to the output with a \textit{null} prompt. For \textit{image editing}, $\text{logit}_c$ refers to the output with both text instruction and image input, while $\text{logit}_u$  refers to the output with only image input (no text instruction).

It is important to note that language AR models do not require logit adjustment, while Selftok does. This is due to resource limitations, as the current VLM is relatively small in scale and has not undergone pretraining on the visual modality, which leads to high entropy at certain timesteps due to dataset bias~\cite{niu2021introspective, mu2025editar}. However, this issue can be mitigated through future large-scale visual pretraining using videos~\cite{liu2024world}.

\begin{table}[t]
\centering
\resizebox{\linewidth}{!}{
\begin{tabular}{l|c|c|cccc|cc}
\toprule
\multirow{2}{*}{Method} & \multirow{2}{*}{T2I Model} & \multirow{2}{*}{\begin{tabular}[c]{@{}c@{}}Structure \\ Distance ↓\end{tabular}} & \multicolumn{4}{c|}{Background Preservation}                                                   & \multicolumn{2}{c}{CLIP Similarity}  \\ \cline{4-9} 
                        &                            &                                                                                 & \multicolumn{1}{c|}{PSNR ↑} & \multicolumn{1}{c|}{LPIPS ↓} & \multicolumn{1}{c|}{MSE ↓}   & SSIM ↑ & \multicolumn{1}{c|}{Whole ↑} & Edited ↑ \\ \hline
Prompt-to-Prompt \cite{hertz2022prompt}       & SD1.4                      & 69.43                                                                           & \multicolumn{1}{c|}{17.87} & \multicolumn{1}{c|}{208.80}  & \multicolumn{1}{c|}{219.88} & 71.14 & \multicolumn{1}{c|}{25.01}  & 22.44   \\ 
Null-text Inversion \cite{mokady2023null}    & SD1.4                      & 13.44                                                                           & \multicolumn{1}{c|}{27.03} & \multicolumn{1}{c|}{60.67}  & \multicolumn{1}{c|}{35.86}  & 84.11 & \multicolumn{1}{c|}{24.75}  & 21.86   \\ 
PnP Inversion \cite{ju2023direct}          & SD1.4                      & 11.65                                                                           & \multicolumn{1}{c|}{27.22} & \multicolumn{1}{c|}{54.55}  & \multicolumn{1}{c|}{32.86}  & 84.76 & \multicolumn{1}{c|}{25.02}  & 22.10    \\ 
Pix2Pix-Zero \cite{parmar2023zero}          & SD1.4                      & 61.68                                                                           & \multicolumn{1}{c|}{20.44} & \multicolumn{1}{c|}{172.22} & \multicolumn{1}{c|}{144.12} & 74.67 & \multicolumn{1}{c|}{22.80}   & 20.54   \\ 
MasaCtrl \cite{cao2023masactrl}               & SD1.4                      & 28.38                                                                           & \multicolumn{1}{c|}{22.17} & \multicolumn{1}{c|}{106.62} & \multicolumn{1}{c|}{86.97}  & 79.67 & \multicolumn{1}{c|}{23.96}  & 21.16   \\ \hline
InstructPix2Pix \cite{brooks2023instructpix2pix}        & SD1.5                      & 107.43                                                                          & \multicolumn{1}{c|}{16.69} & \multicolumn{1}{c|}{271.33} & \multicolumn{1}{c|}{392.22} & 68.39 & \multicolumn{1}{c|}{23.49}  & 22.20    \\ 
MagicBrush \cite{zhang2023magicbrush}             & SD1.5                      & 26.81                                                                           & \multicolumn{1}{c|}{26.85} & \multicolumn{1}{c|}{66.67}  & \multicolumn{1}{c|}{171.11} & 83.37 & \multicolumn{1}{c|}{23.89}  & 20.84   \\ 
InstructDiffusion \cite{geng2024instructdiffusion}      & SD1.5                      & 74.21                                                                           & \multicolumn{1}{c|}{20.88} & \multicolumn{1}{c|}{142.35} & \multicolumn{1}{c|}{353.45} & 76.70  & \multicolumn{1}{c|}{24.06}  & 21.57   \\
MGIE \cite{fu2023guiding}                   & SD1.5                      & 67.41                                                                           & \multicolumn{1}{c|}{21.20}  & \multicolumn{1}{c|}{142.25} & \multicolumn{1}{c|}{295.11} & 77.52 & \multicolumn{1}{c|}{24.28}  & 21.79   \\ 
SEED-X-Edit \cite{ge2024seed}            & SD-XL                      & 61.69                                                                           & \multicolumn{1}{c|}{18.80}  & \multicolumn{1}{c|}{173.63} & \multicolumn{1}{c|}{209.05} & 74.93 & \multicolumn{1}{c|}{25.51}  & 22.20    \\ 
EditAR \cite{mu2025editar}                 & LlamaGen                   & 39.43                                                                           & \multicolumn{1}{c|}{21.32} & \multicolumn{1}{c|}{117.15} & \multicolumn{1}{c|}{130.27} & 75.13 & \multicolumn{1}{c|}{24.87}  & 21.87   \\ 
Selftok (Ours)                    & Llama3.1                     & 35.89                                                                           & \multicolumn{1}{c|}{23.76} & \multicolumn{1}{c|}{124.48} & \multicolumn{1}{c|}{76.78}  & 78.46 & \multicolumn{1}{c|}{24.94}  & 22.02   \\ \bottomrule
\end{tabular}
}
\caption{Quantitative comparison on PIE-Bench dataset \cite{ju2023direct} with various single-image inversion-based methods (top) and large-scale training approaches (bottom). Among the large-scale training ones, EditAR \cite{mu2025editar} is an autoregressive framework.}
\label{tab:edit}
\end{table}

\subsubsection{Text-to-Image Generation}

\noindent\textbf{Metric Evaluation.}
To evaluate the performance of our model, we conduct tests on two widely-used benchmarks: GenEval~\cite{ghosh2023geneval} and DPG-Bench~\cite{DPG-Bench}.
The evaluation results are shown in Tables~\ref{tab:main_geneval} and~\ref{tab:main_DPG}, respectively, which incorporate both stages of the model training: cross-modality training (Selftok-Pre) and cross-task training (Selftok-SFT). On the GenEval benchmark, Selftok-Pre achieves a score of 0.60, outperforming Emu3-Gen (0.54) and TokenFlow (0.55). After SFT, the score of our model increases to 0.74, surpassing Infinity (0.73) and CogView4 (0.73). Similarly, Selftok-SFT reaches 81.8, exceeding Emu3-Gen (80.6) on the DPG benchmark. This highlights that our model enables better image-text alignment.

\noindent\textbf{Qualitative Examples.}
In Figure~\ref{fig:t2i_example}, we visualize the performance of Selftok on the DPG test prompt. We also compare our model with MidJourney and FLUX~\cite{flux2024}, showing that Selftok performs well in both adhering to complex semantics and generating aesthetically pleasing images. However, it should be noted that the current model can only generate images at a resolution of $256 \times 256$, indicating significant potential for improvement in image detail in future work.

\begin{figure}[!htp]
    \centering
    \includegraphics[width=\textwidth]{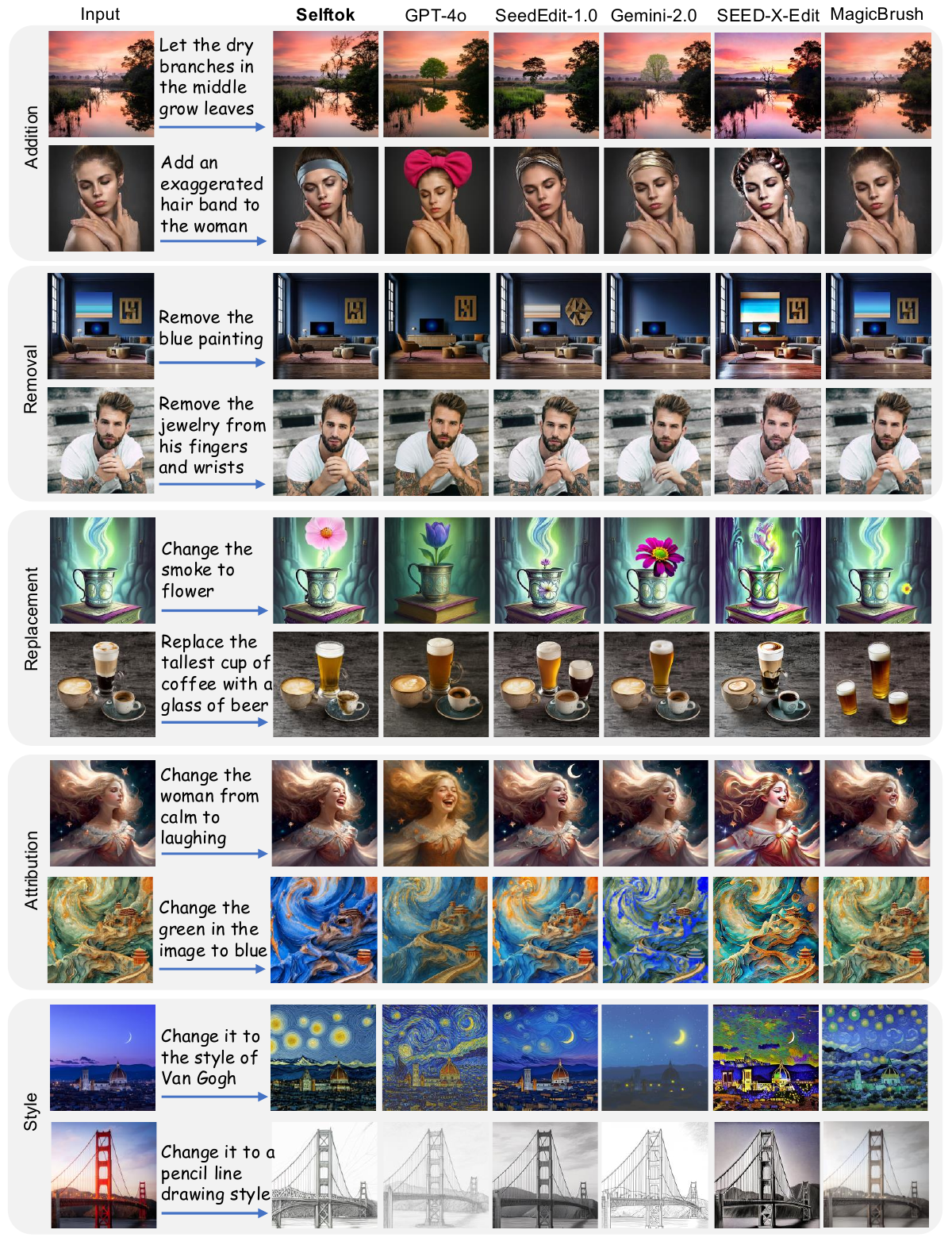}
    \caption{Qualitative comparison on single-turn editing across five editing types.}
    \label{fig:single_editing}
\end{figure}
\begin{figure}[!htp]
    \centering
    \includegraphics[width=\textwidth]{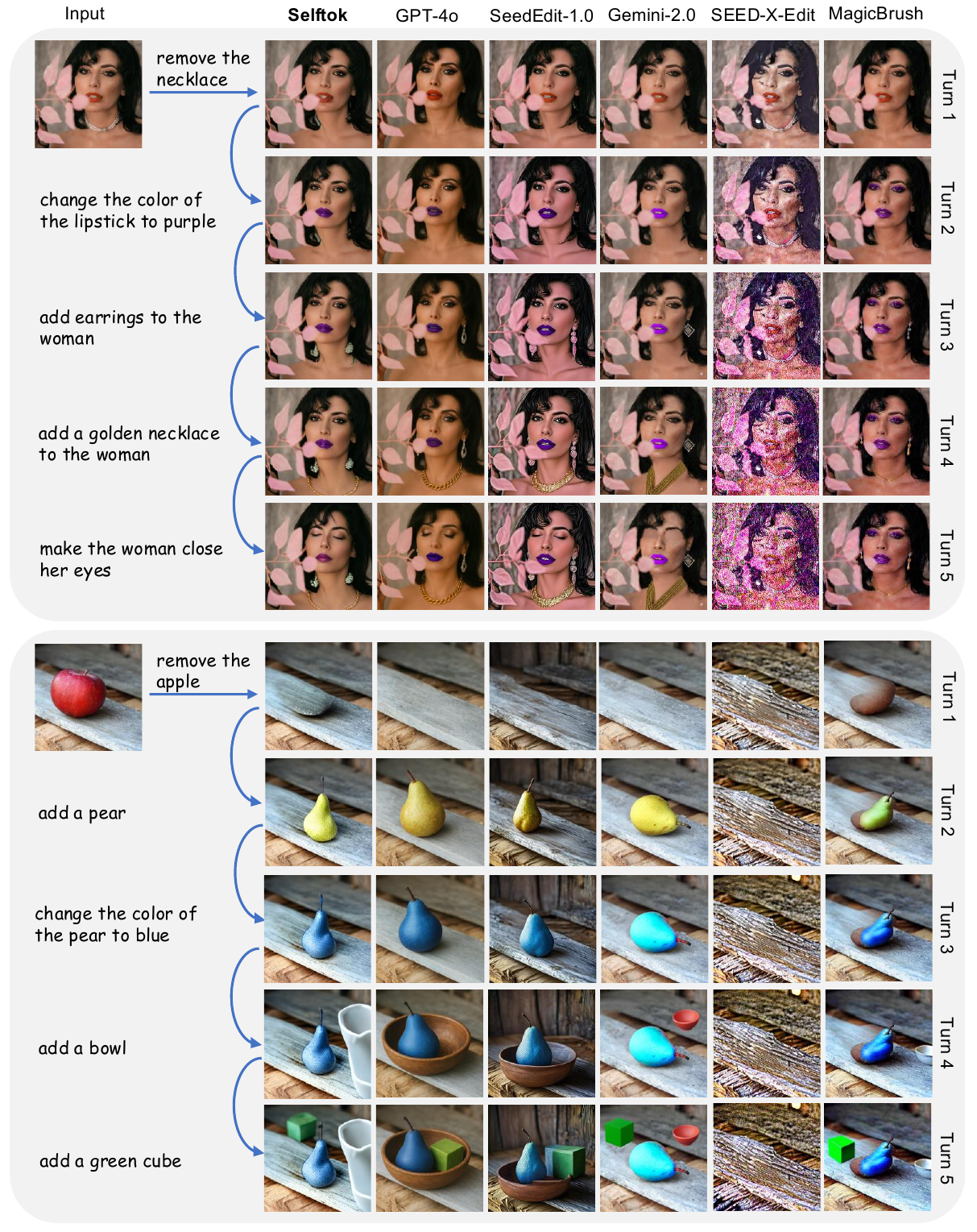}
    \caption{Qualitative comparison on multi-turn editing.}
    \label{fig:multi_editing}
\end{figure}

\subsubsection{Image Editing}

\noindent\textbf{Metric Evaluation.} In Table~\ref{tab:edit}, we conduct a quantitative comparison with previous single-image inversion-based methods and large-scale training ones. Specifically, Structure Distance and Background Preservation metrics reflect fidelity while CLIP Similarity is adopted for evaluating editability. Among the inversion methods, optimization-based approaches, \ie, Null-text Inversion~\cite{mokady2023null} and PnP Inversion~\cite{ju2023direct} achieve relatively better fidelity and editability. However, they are time-consuming during test time. For large-scale training approaches, MagicBrush~\cite{zhang2023magicbrush} obtains the best fidelity but the worst editability. By contrast, SEED-X-Edit~\cite{ge2024seed} exhibits the best performance in editability and is worse in fidelity. They fail to achieve a good balance between these two. Served as an autoregressive framework, EditAR~\cite{mu2025editar} shows a better trade-off. In comparison with EditAR~\cite{mu2025editar}, our Selftok excels in most metrics and obtains the best results on the PIE-Bench dataset \cite{ju2023direct}. It is worth noting that a standardized and objectively reliable evaluation metric for editing has yet to be established in our community. As a result, human evaluation remains the \textit{de facto} ``gold'' standard.

\begin{figure*}
    \centering
    \footnotesize
    \includegraphics[width=\textwidth]{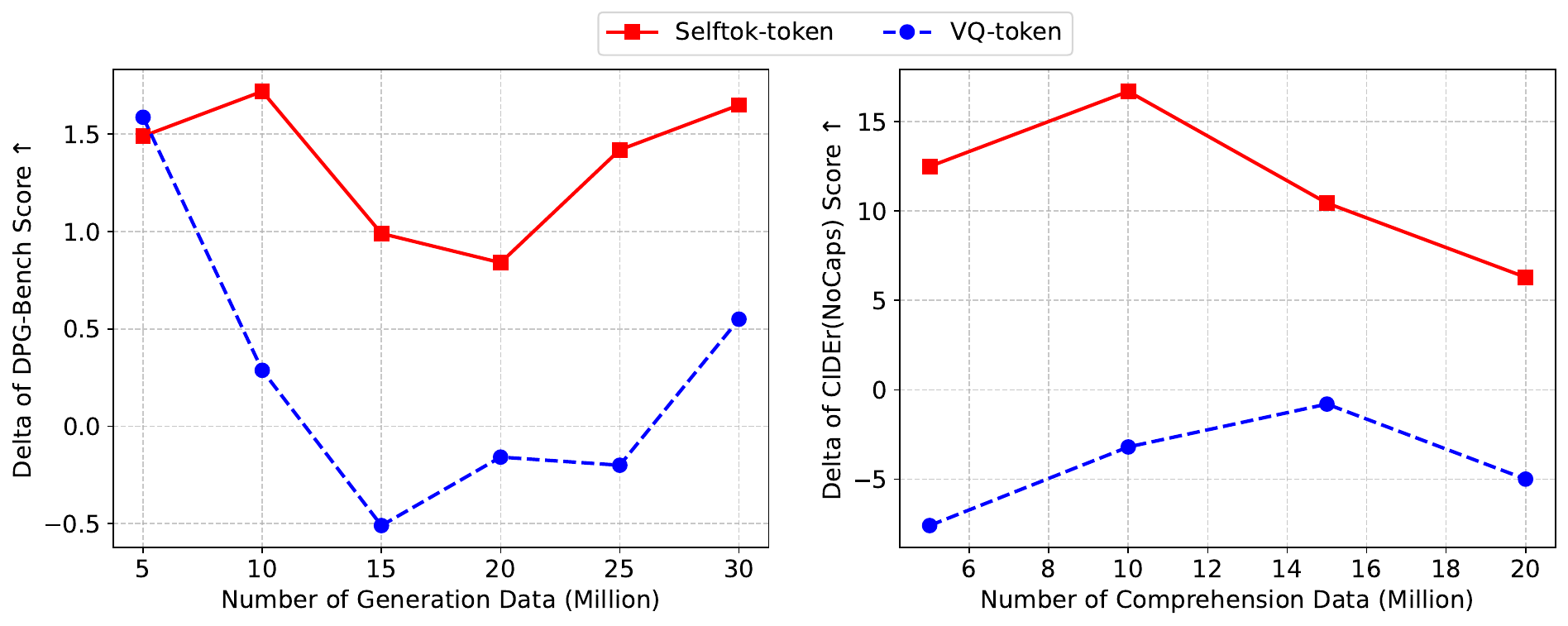}
    \vspace*{-5mm}
    \caption{
    Synergy comparison between Selftok and spatial VQ-tokens. $\Delta$ is computed as the performance gain from dual-task training over single-task training. Left: Generation task performance (DPG-Bench). Right: Comprehension task performance (NoCaps). 
    }

    \vspace*{-1mm}
    \label{fig:synergy}
\end{figure*}

\noindent\textbf{Qualitative Examples.} To have a comprehensive evaluation of our Selftok, we perform a qualitative comparison with leading works, \ie, GPT-4o, SeedEdit-1.0~\cite{shi2024seededit}, Gemini-2.0, SEED-X-Edit~\cite{ge2024seed}, and MagicBrush \cite{zhang2023magicbrush}, on both single-turn and multi-turn editing. For the comparison of single-turn editing across five editing types in Figure~\ref{fig:single_editing}, GPT-4o, SeedEdit-1.0~\cite{shi2024seededit}, and Gemini-2.0 achieve better performance considering both the fidelity and editability. It could be seen that our Selftok gets comparative results with these three. Figure~\ref{fig:multi_editing} gives two examples of multi-turn editing with five turns. Among the compared methods, GPT-4o exhibits the strongest editing capability in the multi-turn setting. Our Selftok demonstrates relatively better performance in terms of text-image alignment and identity preservation across all turns.

\subsubsection{Vision-Language Comprehension}
\noindent\textbf{Metric Evaluation.} We evaluate the models on the visual comprehension benchmark MME, comparing two main categories of methods: ``Specialized Visual Understanding MLLMs'' and ``MLLMs for both Visual Understanding and Generation''. As shown in Table~\ref{tab:mllm}, Selftok achieves a score of 1381.3, outperforming methods that jointly perform generation and understanding (\eg, Emu3 with 1292.8 and VILA-U with 1338.1). However, it still falls short of models specifically designed for comprehension tasks, such as LLaVA (1532.1); or those equipped with an additional comprehension branch, such as Janus-Pro (1576.0). This discrepancy can be attributed to the lack of specialized data for training on comprehension tasks. Recent studies have demonstrated the effectiveness of RL in enhancing vision-language comprehension tasks~\cite{shen2025vlm,peng2025skywork,huang2025vision}. Recall that comprehension tasks produce only language outputs; therefore, these RL methods can be regarded as natural extensions of the success of RL in LLMs and thus fall outside the scope of this work. In contrast, we argue that RL for visual generation remains largely underexplored and presents unique challenges, which we examine in greater detail in Section~\ref{sec:4}.

\begin{table}[h]
\centering
\resizebox{\textwidth}{!}{
\begin{tabular}{@{}llc|ccccccccccc@{}}
\toprule
\textbf{Type} & \textbf{Method} & \textbf{\#Params} & \textbf{Exist.} & \textbf{Count} & \textbf{Pos.} & \textbf{Color} & \textbf{Poster} & \textbf{Celeb.} & \textbf{Scene} & \textbf{Landm.} & \textbf{Art.} & \textbf{OCR} & \textbf{Total} \\
\midrule
\multirow{5}{*}{\textit{Und. Only}}
& InstructBLIP~\cite{dai2023instructblip} & 13B          & 185.0 & 143.3 & 66.7  & 153.3 & 123.8 & 101.2 & 153.0 & 79.8  & 134.3 & 72.5 & 1212.8 \\
& Qwen-VL-Chat~\cite{Qwen-VL}   & 7B & 158.3 & 150.0 & 128.3 & 170.0 & 178.6 & 120.6 & 152.3 & 164.0 & 125.5 & 140.0 & 1487.6  \\
& LLaVA-v1.5~\cite{liu2024improved}         & 7B         & 185.0 & 155.0 & 133.3 & 170.0 & 160.5 & 152.9 & 161.2 & 170.5 & 118.8 & 125.0 & 1532.1 \\
& InternVL-Chat-V1.5~\cite{chen2024internvl} & 20B & 190.0 & 	175.0 & 166.7 & 178.3 & 173.8 & 138.5 & 154.8 & 177.8 & 143.0 & 140.0 & 1637.8\\
& InternLM-XComposer2-VL~\cite{internlmxcomposer2} & 7B & 195.0 & 160.0 & 163.3 & 195.0 & 171.1 & 153.8 & 164.8 & 176.0 & 185.5 & 147.5 & 1712.0 \\ 
\midrule
\multirow{6}{*}{\textit{Und. \& Gen.}}
& Show-o~\cite{xie2024show}            & 1.3B          & 190.0 & 103.3 & 106.7 & 170.0 & 46.9  & 53.8  & 156.5 & 116.5 & 102.0 & 57.5 & 1103.3 \\
& Emu3~\cite{wang2024emu3} & 8B                & \textbf{200.0}          & 145.0        & \textbf{146.7}         & 170.0          & 109.5           & 79.4            & 157.2          & 121.0           & 94.0          & 70.0         & 1292.8         \\
& VILA-U~\cite{wu2024vila}           & 7B  & 165.0 & 136.7 & 121.7 & 158.3 & \textbf{127.9} & 106.8 & 160.2 & 151.2 & 122.8 & 87.5 & 1338.1 \\
& Liquid~\cite{wu2025liquid} & 7B                & 163.3           & 96.7           & 115.0           & 133.3          & 75.5            & 84.1            & 140.0            & 82.5            & 86.0            & 55.0           & 1031.5         \\
& Janus-Pro~\cite{chen2025janus} & 7B & 185.0          & \textbf{161.7}   & 128.3   & 175.0         & 152.4          & \textbf{159.4}    & \textbf{163.5}    & 157.8      & \textbf{123.0}   & \textbf{170.0}     & \textbf{1576.0} \\
\cmidrule{2-14}
& \textbf{Selftok} & 8B                & 180.0           & 116.7          & 118.3         & \textbf{175.0  }        & 122.8           & 121.5           & 161.5        & \textbf{176.8}           & 91.3          & 117.5        & 1381.3 \\
\bottomrule
\end{tabular}
}
\caption{Comparison for multimodal comprehension on MME vision-language benchmarks. 
All the unified model scores are reproduced using the official open-source code, as the original papers do not provide scores for individual components.
}
\label{tab:mllm}
\end{table}


\noindent\textbf{Synergy between Generation and Comprehension}: The synergy of visual comprehension and generation is challenging because of the conflicting objectives~\cite{pan2024auto}:  for comprehension, the VLM needs to abstract the visuals (invariance); for generation, it needs to preserve the visuals as much as possible (equivariance). We conduct ablation experiments and introduce the following metric:
\begin{equation}
\Delta = \text{Performance}_{\text{(Both Tasks)}} - \text{Performance}_{\text{(Single Task)}},
\end{equation}
where ``Both Tasks'' refers to training on understanding and generation with a fixed data ratio (\ie, 4:6), and ``Single Task'' refers to the generation or the comprehension, as shown in Figure~\ref{fig:synergy}. We visualize the results for different amounts of generation data to mitigate the risk of underfitting. The results demonstrate that Selftok consistently outperforms spatial tokens (the VQToken adopted in~\cite{chen2025janus}), with \(\Delta\) remaining positive across all data levels, highlighting a clear synergy between understanding and generation tasks. In contrast, the spatial tokens exhibit negative \(\Delta\) values at certain points, suggesting a conflict. Note that this synergy holds great potential for post-training in vision-language models (VLMs), as comprehension must be used as the reward model to guide reinforcement learning for visual generation (Section~\ref{sec:4.2.1}).

\section{Selftok-based Visual RL}
\label{sec:4}

There is no universally ``correct'' ground-truth label for multimodal alignment. For instance, a single text prompt in text-to-image generation may correspond to infinitely many plausible images. Consequently, a VLM trained solely on vision-language pairs tends to naively mimic the training distribution rather than perform genuine reasoning, resulting in the well-known hallucination issue~\cite{lu2025towards,li2025treble,vice2025exploring}.
This is shown in Figure~\ref{fig:results_rl}, where the Selftok-SFT model cannot generate ``yellow broccoli'' or ``red dog'' due to the rare combinations in training data (see Appendix~\ref{sec:app_hal} for more details).
A straightforward remedy is to curate a supervised dataset that exhaustively covers all possible alignments. However, this approach is unsustainable.

This motivates us to the need for \textit{visual reinforcement learning (visual RL)}, a paradigm that visual generative models are trained \textit{without} ground-truth images, relying instead on task-specific rewards, such as the CLIP-based image-prompt similarity~\cite{radford2021learning} for text-to-image generation. Inspired by the success of RL in LLMs, which unlocks their reasoning capabilities, we believe that visual RL can similarly advance visual generation from mere imitation to genuine reasoning.

Next, we formulate the problem of visual RL in Section~\ref{sec:4.1}, which reveals that AR tokens are necessary in this setup.
In Section~\ref{sec:4.2}, we design an RL training pipeline for Selftok.
Thanks to its AR property, we show in Section~\ref{sec:4.3} that this leads to significant improvements in GenEval and DPG-Bench, which are not yet observed for non-AR spatial tokens.

\subsection{Problem Formulation}
\label{sec:4.1}
\begin{wrapfigure}{r}{0.35\textwidth}
\vspace{-10mm}
\footnotesize
\includegraphics[width=\linewidth]{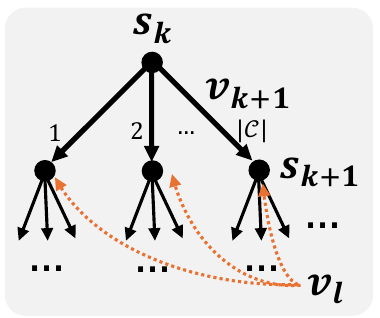}
\vspace{-5mm}
\caption{\justifying The recursive Bellman equation fails when a child node $v_{l}$ (\ie, future token) anti-causally affects a parent node $v_{k+1}$.}
\label{fig:8}
\vspace{-10mm}
\end{wrapfigure}

In visual RL, we aim to fine-tune a VLM (policy) that selects the next token (action) based on the current sequence (state) to maximize a task-specific reward (\eg, the consistency between the text prompt and generated image). Without loss of generality, we limit our discussion to visual tokens $[v_1,\ldots,v_K]$, as the same principle applies to language tokens.
We discuss the recipe for visual RL in detail:
\\
\textbf{1) State}: The state $s_k=[v_1,\ldots,v_k]$ is the token sequence generated by VLM at step $k\in \{1,\ldots, K\}$, and the initial state $s_0=[]$ is defined as an empty sequence.
\\
\textbf{2) Action}: An action at step $k$ selects the next token $v_{k+1}$ from the visual codebook $\mathcal{C}$, \ie, at each step, there are $|\mathcal{C}|$ possible actions to choose from.
\\
\textbf{3) State transition}: $P(s_{k+1} | s_k, v_{k+1})=1$ because $s_{k+1}=[s_k, v_{k+1}]$.
\\
\textbf{4) Reward}: Generally, the reward $r(s_{k},v_{k+1})$ received at step $k+1$ depends on the previous state $s_{k}$ (where the reward is from) and the action as the next token $v_{k+1}$, predicted at the previous state (how the reward is obtained). With the state and action defined above, $s_{k+1} = [s_k, v_{k+1}]$, we can also write $r(s_{k+1})=r(s_{k},v_{k+1})$.
\\ 
\textbf{5) Policy}: Given the current state $s_k$, the policy $\pi(v_{k+1}|s_k)$ predicts an action as the next token $v_{k+1}$. The goal of RL is to find an optimal policy $\pi$, which generates a trajectory  $s_0\rightsquigarrow s_K$ that maximizes
the cumulative reward (with omitted discount factor):
\begin{equation}
    \mathop{\mathrm{max}}_\pi V_\pi(s_0), \quad \text{where} \quad V_\pi (s_k) = \mathbb{E}_\pi \left[ \sum_{i=k}^{K-1} r(s_{i},v_{i+1}) \right].
    \label{eq:rl}
\end{equation}
$V_\pi(s_k)$ is the value function, accounting for the expectation of all the possible cumulative rewards received along the trajectory $s_k\rightsquigarrow s_K$ generated by $\pi$.

We now show that only AR tokens can derive the Bellman equation, which underpins the optimality of policy update that guarantees effective RL. We start by rewriting our goal $V_\pi(s_0)$ in Eq.~\eqref{eq:rl}:
\begin{align}
    V_\pi(s_0) &= \mathop{\mathbb{E}}_{[v_1\sim \pi(\cdot|s_0), v_2\sim \pi(\cdot|s_1), \ldots, v_{K}\sim \pi(\cdot|s_{K-1})]} \left[r(s_0, v_1)+r(s_1,v_2)+ \ldots+ r(s_{K-1},v_K)\right] \label{eq:bellman_1}\\
    &= \mathop{\mathbb{E}}_{v_1\sim \pi(\cdot|s_0)} r(s_0,v_1) + \mathop{\mathbb{E}}_{v_1\sim \pi(\cdot|s_0)}\underbrace{\mathop{\mathbb{E}}_{[v_2\sim \pi(\cdot|s_1), \ldots, v_{K}\sim \pi(\cdot|s_{K-1})]}\left[r(s_1,v_2)+ \ldots+ r(s_{K-1},v_K)\right]}_{V_\pi(s_1)} \label{eq:bellman_2}\\
    &= \sum_{v_{1} \in \mathcal{C}} \pi(v_{1}|s_0) \cdot \left[ r(s_{1}) + V_\pi(s_{1}) \right].
    \label{eq:bellman_3}
\end{align}
Eq.~\eqref{eq:bellman_1} holds because the transition probability $P(s_{k+1} | s_k, v_{k+1})=1$.  As shown in Figure~\ref{fig:8},  Eq.~\eqref{eq:bellman_2} holds because of the causal dependency of AR, where the choice of action $v_{k+1}$ only depends on $s_k$ and does not affect the former action $v_{k}$ that has already been chosen. Therefore, we can recursively apply Eq.~\eqref{eq:bellman_3} and derive the Bellman equation:
\begin{equation}
    V_\pi(s_k) = \sum_{v_{k+1} \in \mathcal{C}} \pi(v_{k+1}|s_k) \cdot \left[ r(s_{k+1}) + V_\pi(s_{k+1}) \right].
\end{equation}
Thanks to the above equation, the optimized $\pi$ in Eq.~\eqref{eq:rl} can be step-by-step obtained: 
\begin{equation}
\mathop{\mathrm{argmax}}_{v_{k+1}} \pi'(v_{k+1}|s_{k}) \leftarrow     \mathop{\mathrm{argmax}}_{v_{k+1}}\left[r(s_{k+1}) +  V_\pi(s_{k+1})\right].
\label{eq:policy}
\end{equation}
Although the above policy update is greedy, its optimality is guaranteed by the policy improvement
theorem~\cite{sutton1998reinforcement}, which shows that the locally optimal action $v_{k+1}$ at step $k$ does not affect the earlier improved actions due to the AR property. Note that non-AR spatial tokens cannot satisfy the Bellman equation, and therefore cannot support the policy update that relies on it. The key reason is that Eq.~\eqref{eq:bellman_2} cannot be derived, as the future action $v_l$, where $l>k+1$, influences earlier actions through the anti-causal links (shown red in Figure~\ref{fig:8}). Therefore, spatial tokens are not compatible with RL.

\subsection{Implementation}
\label{sec:4.2}
We now describe the implementation details for Selftok-based visual RL for visual generation, such as text-to-image and visual editing tasks, \textbf{without using any pairwise supervision}, including two reward models for evaluating the quality of the generated images and training objectives for updating the policy network.
\subsubsection{Reward Model}
\label{sec:4.2.1}

The overall design philosophy of our reward model is to utilize visual comprehension models to evaluate the generated image in visual RL and provide feedback for optimization. For tasks such as text-to-image generation, the comprehension model should understand and evaluate the consistency between the generated image and the textual prompt. In this paper, we categorize the comprehension-based reward into two major types:
\\
\textbf{Program-based Reward:} This type is useful for more structured tasks like object identification, counting, and spatial relationships~\cite{ghosh2023geneval}, where the prompt explicitly and unambiguously states the desired generation, \eg, ``\textit{3 clocks and 1 dog}'', and thus we can use visual detectors~\cite{chen2019mmdetection} to evaluate the generation quality. For example, we count the clocks based on the detector’s confidence, returning 1 if the count is correct and 0 otherwise.
Each prompt has its own item sets to be tested, and the average of the scores for each test is used as the reward score.
\\
\textbf{QA-based Reward:} For more complex and ambiguous prompts, it is challenging to rely solely on automated programs. To this end, we resort to more powerful visual comprehension models like InternVL~\cite{chen2024internvl} or GPT-4o~\cite{hurst2024gpt}, which can comprehend nuanced prompts and generate accurate answers. 
Specifically, inspired by~\cite{cho2023davidsonian}, we first decompose the prompt to semantic tuples (\eg, entity, attribute, and relation) and then generate questions (\eg, \textit{``Is the car red?''}). The MLLMs are asked to perform a VQA task for the prompt and generated image, returning a score of 0 to 1 (\eg, wrong to correct) for each question. The reward is obtained by averaging the evaluation of the MLLMs on multiple questions for a prompt.
We can also fine-tune such models to obtain more task-specific reward functions.

As a preliminary study, we only validate the feasibility of the above two types. However, we believe that there should be more effective comprehension tasks as reward models for better performance, and we leave the exploration of them for future work.

\subsubsection{Policy Gradient}
We adopt a simplified version of GRPO~\cite{shao2024deepseekmath} without importance sampling and encourage readers to explore more advanced alternatives.
For each prompt, the policy network $\pi$ generates a batch of outputs $\{s^i\}_{i=1}^B$, where $B$ represents the batch size and each $s^i$ denotes the final state $[v_0,v_1,\ldots,v_K]$ of the $i$-th visual sequence. For a batch, we calculate the \textit{total rewards} $\{r(s^i)\}_{i=1}^B$, where we slightly abuse the notation that the total reward $r(s^i) = r(s_K^i)$ as all the intermediate rewards $r(s_k^i) = 0$, $\forall k < K$. We also calculate the advantages $\{A_i\}_{i=1}^B$, where each $A_i$ measures the relative quality of output $s^i$ compared to the average reward:
\begin{equation}
    A_i = \frac{r(s^i) - \text{mean}(\{r(s^1),r(s^2),\ldots,r(s^B)\})}{\text{std}(\{r(s^1),r(s^2),\ldots,r(s^B))},
\end{equation}
where $\text{mean}(\cdot)$ and $\text{std}(\cdot)$ are the mean and standard deviation of all rewards, respectively.

Then, we update the policy network parameters by the following training loss:
\begin{equation}
\mathcal{L} = - \frac{1}{B} \sum_{i=1}^{B}\left[A_i - \lambda \mathbb{D}_{KL}(\pi || \pi_{\text{old}})\right],
\end{equation}
where the KL divergence $\mathbb{D}_{KL}(\pi || \pi_{\text{old}})= \frac{\pi_{\text{old}}}{\pi}-\text{log}\frac{\pi_{\text{old}}}{\pi}-1$ is to maintain training stability.
It measures the difference between the new policy $\pi$ and the old policy $\pi_{\text{old}}$, where the new policy $\pi$ is the up-to-date one after policy gradient; the old policy $\pi_{\text{old}}$ refers to the one used to generate the token sequences before the policy gradient update.

\noindent\textbf{Difference from diffusion-based DPO}. Direct Preference Optimization (DPO) in diffusion models~\cite{wallace2024diffusion} has long been mistakenly regarded as a visual RL method. Compared to our Selftok-based visual RL---which aligns with the standard formulation of RL in LLMs---the diffusion-based DPO essentially degenerates to contrastive learning between the ``good'' and ``bad'' explored samples. This limitation arises because diffusion models do not have access to the actual\footnote[4]{As the diffusion inversion is notoriously hard~\cite{mokady2023null}, they instead resort to the forward pass to sample ``easy'' but inaccurate trajectories.} generation trajectories of the samples, \ie, they lack the state-action trajectories available in the original DPO formulation for LLMs~\cite{rafailov2023direct}. In contrast, Selftok discretizes the entire diffusion generation path of an image, thereby encoding a complete state-action trajectory.

\subsection{Validation}
\label{sec:4.3}
In this section, we experimentally evaluate the text-to-image generation capabilities of the Selftok-based AR model, demonstrating the effectiveness of visual RL. We also provide details of the visual RL training and analyze the impact of various factors on the model performance.

\subsubsection{Implementation details}
Based on the SFT model (see Section~\ref{sec:3}), we further apply the reward model (see Section~\ref{sec:4.2.1}) for visual RL and evaluate its performance on Geneval~\cite{ghosh2023geneval} and DPG-Bench~\cite{hu2024ella}. 

For program-based reward, we use MM-Detection~\cite{chen2019mmdetection} as the detectors and set the threshold for detection to 0.6. 
For QA-based reward, we utilize InternVL~\cite{chen2024internvl} and mPLUG~\cite{li2022mplug} as the comprehension model.
Note that we carefully deduplicate the training prompts to ensure that there is no overlap with the test set. For the sake of reproducibility, after the visual RL training, \textbf{we do not incorporate any test-time scaling techniques during inference.}

\subsubsection{Main Results}
\label{sec:4.3.1}

The quantitative experimental results are summarized in Table~\ref{tab:main_geneval} and Table~\ref{tab:main_DPG}, which evaluate the performance of our Selftok-based approach on the GenEval and DPG-Bench benchmarks. 
Specifically, Selftok-Pre and Selftok-SFT refer to the models after stage 1 (cross-modality alignment) and stage 2 (cross-task alignment), respectively (see Section~\ref{sec:3.1}), while Selftok-Zero refers to the model after visual reinforcement learning (see Section~\ref{sec:4.2}).

\begin{figure*}
    \centering
    \footnotesize
    \vspace*{-8mm}
    \begin{subfigure}[t]{\textwidth}
         \includegraphics[width=\textwidth]{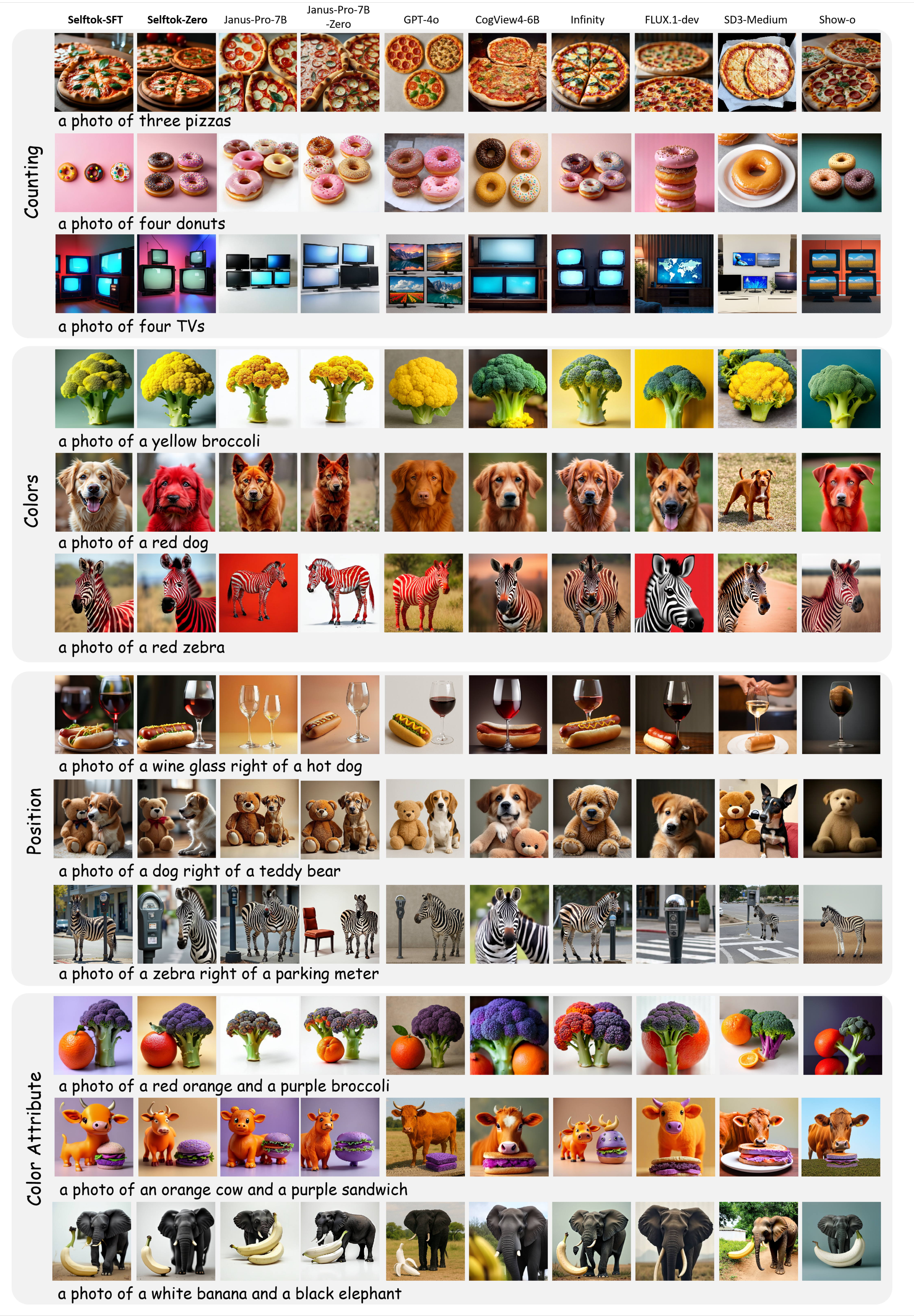}
         \phantomcaption
    \end{subfigure}
    \vspace*{-4mm}
    \caption{Qualitative experimental results of Selftok-based visual RL. Compared to existing text-to-image generation models, the images generated by Selftok demonstrate better alignment with the given prompts.}
    \vspace*{-5mm}
    \label{fig:results_rl}
\end{figure*}

\noindent\textbf{Selftok-Zero achieves state-of-the-art performance in text-to-image generation.}
As shown in Table~\ref{tab:main_geneval}, Selftok-Zero obtains the highest overall score of \underline{95 on the GenEval} benchmark, surpassing all previous models, including strong baselines such as CogView4-6B (73) and HiDream-I1 (83). Selftok-Zero also outperforms across all major sub-tasks, \eg, \underline{Colors (97)} and \underline{Position (98)}. 
Similarly, on DPG-Bench (Table~\ref{tab:main_DPG}), Selftok-Zero achieves an overall score of \textbf{85.57}, outperforming SD3-Medium (84.08) and Janus-Pro-7B (84.19).
The qualitative results are presented in Figure~\ref{fig:results_rl}, the images generated by Selftok-Zero exhibit high-quality alignment with the textual descriptions.

\noindent\textbf{Visual RL significantly enhances image-text consistency.}
A direct comparison of Selftok-SFT \textit{vs} Selftok-Zero and Janus-Pro-7B$^\dagger$ \textit{vs} Janus-Pro-7B-Zero highlights the benefits of visual RL. On GenEval, Selftok-Zero improves upon its supervised counterpart in nearly every metric, with notable gains in \underline{Position (45$\rightarrow$98)} and \underline{Counting (66$\rightarrow$90)}. 
On DPG-Bench, visual RL leads to a +3.77 increase in overall score, with improvements in Entity (from 88.15$\rightarrow$91.78) and Relation (from 93.68$\rightarrow$95.26). These results indicate that visual RL is effective in closing the gap between generated images and complex textual prompts.


\begin{figure*}
    \centering
    \footnotesize
    \begin{subfigure}[t]{0.97\textwidth}
         \includegraphics[width=\textwidth]{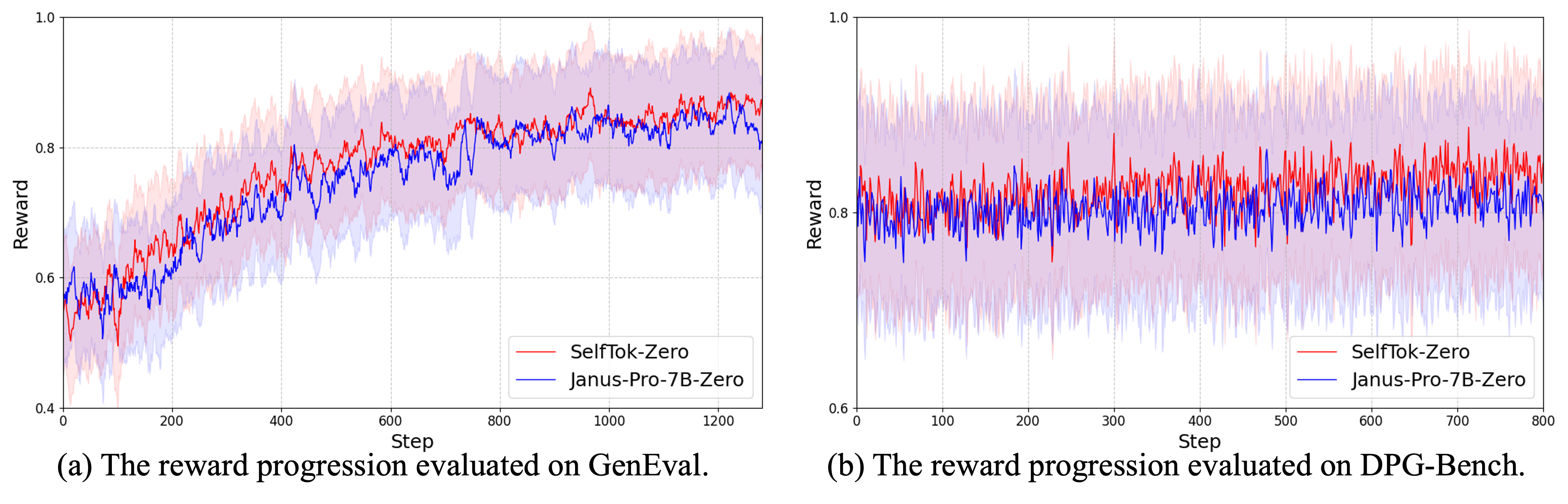}
         \phantomcaption
         \label{fig:rl_reward_all}
    \end{subfigure}
    \begin{subfigure}[t]{0\textwidth} 
         \includegraphics[width=0\textwidth]{example-image-b}
         \phantomcaption
         \label{fig:rl_reward_dpg}
    \end{subfigure}
    \vspace{-3mm}
    \caption{Comparison of reward progression over steps for Selftok (Selftok-Zero) and spatial tokens (Janus-Pro-7B-Zero) on GenEval and DPG-Bench.}
    \label{fig:rl_reward}
    \vspace{-3mm}
\end{figure*}
\noindent\textbf{Selftok is more effective than spatial tokens in visual RL.}
The results in Table~\ref{tab:main_geneval} and Table~\ref{tab:main_DPG} show that Selftok significantly outperforms spatial token-based methods in visual reinforcement learning (\eg, Janus-Pro-7B-Zero \underline{+6} \textit{vs} Selftok-Zero \underline{+18} on GenEval). Figure~\ref{fig:rl_reward} illustrates the reward score changes during visual RL evaluation on GenEval and DPG-Bench. It is evident that although Janus-Pro-7B$^\dagger$ (79) outperforms Selftok-SFT (74) before visual RL, Selftok-Zero comes from behind to surpass Janus-Pro-7B-Zero (\eg, +7 on Geneval), thanks to the AR properties of Selftok (see Section~\ref{sec:4.1}). These results further highlight the significant impact of the image tokenizer design on visual RL.

\noindent\textbf{Program-based reward yields more substantial gains in visual RL.}
We observe that the improvements on GenEval (program-based reward) are more pronounced than on DPG-Bench (QA-based reward). While Selftok-Zero outperforms Selftok-SFT by +21 in overall score on GenEval \underline{(74$\rightarrow$95)}, the improvement on DPG-Bench is slightly smaller (+3.77, 81.80$\rightarrow$85.57). This suggests that program-based reward—enabled by structured detectors and precise matching—provides stronger and more reliable training signals during reinforcement learning, especially for attributes like object counting, color, and spatial layout.

\begin{table*}
\centering
\resizebox{\textwidth}{!}{
\begin{tabular}{@{}lllllllll@{}}
\toprule
\textbf{Type} & \textbf{Method} & \textbf{Single Obj.} & \textbf{Two Obj.} & \textbf{Counting} & \textbf{Colors} & \textbf{Position} & \textbf{Color Attr.} & \textbf{Overall} \\
\midrule
\multirow{7}{*}{\textbf{Diffusion Only}} 
& PixArt-$\alpha$~\cite{chen2023pixart} & 98 & 50 & 44 & 80 & 8 & 7 & 48 \\
 & SDXL~\cite{podell2023sdxl} & 98 & 74 & 39 & 85 & 15 & 23 & 55 \\
 & FLUX.1-dev~\cite{flux2024} &98  &79  &73  &77  &22  &45  &66  \\
 & DALL-E 3~\cite{shi2020improving} & 96 & 87 & 47 & 83 & 43 & 45 & 67 \\
 & SD3-Medium~\cite{esser2024scaling} & 99 & 94 & 72 & 89 & 33 & 60 & 74 \\
 & CogView4-6B~\cite{CogView4-6B} &99 &86 &66 &79 &48 &58 &73\\
 & HiDream-I1~\cite{hidream} &\textbf{100} &\textbf{98} &79 &91 &60 &72 &83\\
\hline
\multirow{4}{*}{\textbf{Hybrid Model}} 
 & SEED-X~\cite{ge2024seed} & 97 & 58 & 26 & 80 & 19 & 14 & 49 \\
 & Transfusion~\cite{zhou2024transfusion} & - & - & - & - & - & - & 63 \\
 & D-DiT~\cite{li2024dual} & 97 & 80 & 54 & 76 & 32 & 50 & 65 \\
 & Show-o~\cite{xie2024show} & 98 & 80 & 66 & 84 & 31 & 50 & 68 \\
 & GPT-4o$\ddag$~\cite{gpt-4o} & 99 &92 & 85 &91 & 75 & 66 & 85\\
\hline
\multirow{10}{*}{\textbf{Pure dAR}} 
& Emu3-Gen~\cite{wang2024emu3} & 98 & 71 & 34 & 81 & 17 & 21 & 54 \\
 & TokenFlow-XL~\cite{qu2024tokenflow} & 95 & 60 & 41 & 81 & 16 & 24 & 55 \\
 & ILLUME+~\cite{huang2025illume_plus} & 99 & 88 & 62 & 84 & 42 & 53 & 72 \\
 & Infinity~\cite{Infinity} &- &85 &- &- &49 &57 &73\\
 & Janus-Pro-7B~\cite{chen2025janus} & 99 & 89 & 59 & 90 & 79 & 66 & 80 \\
\cmidrule{2-9}
 & Janus-Pro-7B$\dagger$ & 98 & 88 & 58 & 88 & 76 & 65 & 79 \\
 & Janus-Pro-7B-Zero
 & 98$_{\textcolor{ForestGreen}{+0}}$
 & 95$_{\textcolor{ForestGreen}{+7}}$
 & 58$_{\textcolor{ForestGreen}{+0}}$ 
 & 89$_{\textcolor{ForestGreen}{+1}}$
 & 90$_{\textcolor{ForestGreen}{+14}}$
 & 81$_{\textcolor{ForestGreen}{+16}}$
 & 85$_{\textcolor{ForestGreen}{+6}}$ \\
 \cmidrule{2-9}
 & Selftok-Pre & 99 & 57 & 58 & 81 & 22 & 43 & 60 \\
 & Sefltok-Pre-Zero 
 & 99 $_{\textcolor{ForestGreen}{+0}}$
 & 94 $_{\textcolor{ForestGreen}{+37}}$
 & 58 $_{\textcolor{ForestGreen}{+0}}$
 & 89 $_{\textcolor{ForestGreen}{+8}}$
 & 89 $_{\textcolor{ForestGreen}{+67}}$
 & 73 $_{\textcolor{ForestGreen}{+30}}$
 & 84 $_{\textcolor{ForestGreen}{+24}}$ \\
 \cmidrule{2-9}
 & Selftok-SFT & \textbf{100} & 79 & 66  & 91 & 45  & 62  & 74 \\
 & Selftok-Zero 
 &\textbf{100}$_{\textcolor{ForestGreen}{+0}}$ 
 &\textbf{98}$_{\textcolor{ForestGreen}{+19}}$  
 &\textbf{90}$_{\textcolor{ForestGreen}{+24}}$  
 &\textbf{97}$_{\textcolor{ForestGreen}{+6}}$  
 &\textbf{98}$_{\textcolor{ForestGreen}{+53}}$  
 &\textbf{90}$_{\textcolor{ForestGreen}{+28}}$  
 &\textbf{95}$_{\textcolor{ForestGreen}{+21}}$ \\
\bottomrule
\end{tabular}
}
\caption{\justifying Evaluation of text-to-image generation ability on GenEval benchmark. 
Janus-Pro-7B$\dagger$ represents the result of our evaluation.
Janus-Pro-7B-Zero represents a model that has undergone the same visual RL process as Selftok-Pre-Zero and Selftok-Zero.
}
\label{tab:main_geneval}
\end{table*}

\begin{table*}
\centering
\resizebox{\textwidth}{!}{
\begin{tabular}{@{}llllllll@{}}
\toprule
\textbf{Type} & \textbf{Method} & \textbf{Global} & \textbf{Entity} & \textbf{Attribute} & \textbf{Relation} & \textbf{Other} & \textbf{Overall} \\
\midrule
\multirow{7}{*}{\textbf{Diffusion Only}} 
 & PixArt-$\alpha$~\cite{chen2023pixart}  & 74.97 & 79.32 & 78.60 & 82.57 & 76.96 & 71.11\\
 & SDXL~\cite{podell2023sdxl} & 83.27 & 82.43 & 80.91 & 86.76& 
 80.41 & 74.65\\
 &DALL-E 3~\cite{shi2020improving} & 90.97 & 89.61 & 88.39 & 90.58 & 89.83 & 83.50 \\
 &SD3-Medium~\cite{esser2024scaling} & 87.90 & 91.01 & 88.83 & 80.70 & 88.68 & 84.08 \\
 & FLUX.1-dev~\cite{flux2024} & 85.80 & 86.79 & 89.98 & 90.04 & 89.90  & 83.79\\
 & CogView4-6B~\cite{CogView4-6B} & 83.85 & 90.35 & \textbf{91.17} & 91.14 & 87.29 & 85.13\\
  & HiDream-I1~\cite{hidream} & 76.44 & 90.22 & 89.48 & 93.74 &
 \textbf{91.83} & \textbf{85.89}\\
\hline
\multirow{1}{*}{\textbf{Hybrid Model}} 
 & Show-o~\cite{xie2024show} &- &- &- &- &- & 67.48 \\
\hline
\multirow{7}{*}{\textbf{Pure dAR}} 
& Emu3-Gen~\cite{wang2024emu3} & 85.21 & 86.68 & 86.84 & 90.22 & 83.15 & 80.60 \\
 &Janus~\cite{wu2024janus} & 82.33 & 87.38 & 87.70 & 85.46 & 86.41 & 79.68 \\
 & Infinity~\cite{Infinity} &\textbf{93.11} &- &- &90.76 &- &83.46\\
 &Janus-Pro-7B~\cite{chen2025janus} & 86.90 & 88.90 & 89.40 & 89.32 & 89.48 & 84.19 \\
 \cmidrule{2-8}
  &Janus-Pro-7B$\dagger$ & 83.59 & 89.74 & 87.51 & 92.94 & 81.20 & 83.48 \\
  & Janus-Pro-7B-Zero 
 & 84.50$_{\textcolor{ForestGreen}{+0.91}}$
 & 90.13$_{\textcolor{ForestGreen}{+0.39}}$
 & 87.29$_{\textcolor{red}{-0.22}}$
 & 93.44$_{\textcolor{ForestGreen}{+0.50}}$
 & 82.40$_{\textcolor{ForestGreen}{+1.20}}$
 & 84.49$_{\textcolor{ForestGreen}{+1.01}}$\\
 \cmidrule{2-8}
 & Selftok-Pre & 87.41 & 87.09 & 88.08 & 87.89 & 87.42 & 80.37 \\
 &Selftok-SFT & 82.07 & 88.15 & 87.69 & 93.68 & 80.40 & 81.80\\
 &Selftok-Zero 
 &83.59$_{\textcolor{ForestGreen}{+1.52}}$
 &\textbf{91.78}$_{\textcolor{ForestGreen}{+3.63}}$  
 &89.04$_{\textcolor{ForestGreen}{+1.35}}$  
 &\textbf{95.26}$_{\textcolor{ForestGreen}{+1.58}}$  
 &82.80$_{\textcolor{ForestGreen}{+2.40}}$  
 &85.57$_{\textcolor{ForestGreen}{+3.77}}$  \\
\bottomrule
\end{tabular}
}
\caption{Performances on DPG-Bench. The methods in this table are all generation-specific models except Show-o, Janus-Pro, and Selftok.}
\label{tab:main_DPG}
\end{table*}

\section{Conclusion, Limitations, and Ongoing Work}
\label{sec:5}
In this paper, we propose Selftok, a non-spatial 1D visual tokenizer that encodes images into discrete tokens conforming to an autoregressive (AR) structure. We argue that this token structure is crucial for multimodal training: 1) AR visual tokens are inherently compatible with language tokens and large language models (LLMs), which are autoregressive by design; and 2) AR visual tokens enable optimal policy improvement for effective visual reinforcement learning (RL). We demonstrate that Selftok’s AR structure emerges from the reverse diffusion process. To the best of our knowledge, this is the first work to unify diffusion and autoregression within a single LLM framework, without introducing additional architectures or training objectives.

The primary limitation does not lie in Selftok itself, but rather in the significantly slower token generation speed of LLMs compared to diffusion models. For instance, when using 512 tokens per frame, generating a one-minute video clip at 24 fps would require generating $512\times 24\times 60 = 737,280$ tokens---posing a substantial throughput challenge. Fortunately, we are optimistic that this issue will be mitigated by introducing spatial-temporal compression, in conjunction with the rapid progress in real-time massive token generation within the LLM community~\cite{gloeckle2024better}. Another limitation of this work stems from the restricted model scale. Due to limited capacity, we have not yet demonstrated Selftok's ability to transfer visual knowledge to language and realize multimodal emergent capabilities. If resources permit, we plan to investigate the scaling laws of multimodal training with Selftok, aiming to validate its potential for cross-modal synergy. 
Next, we highlight our two ongoing works for Selftok:

\begin{figure*}
    \centering
    \footnotesize
    \includegraphics[width=\textwidth]{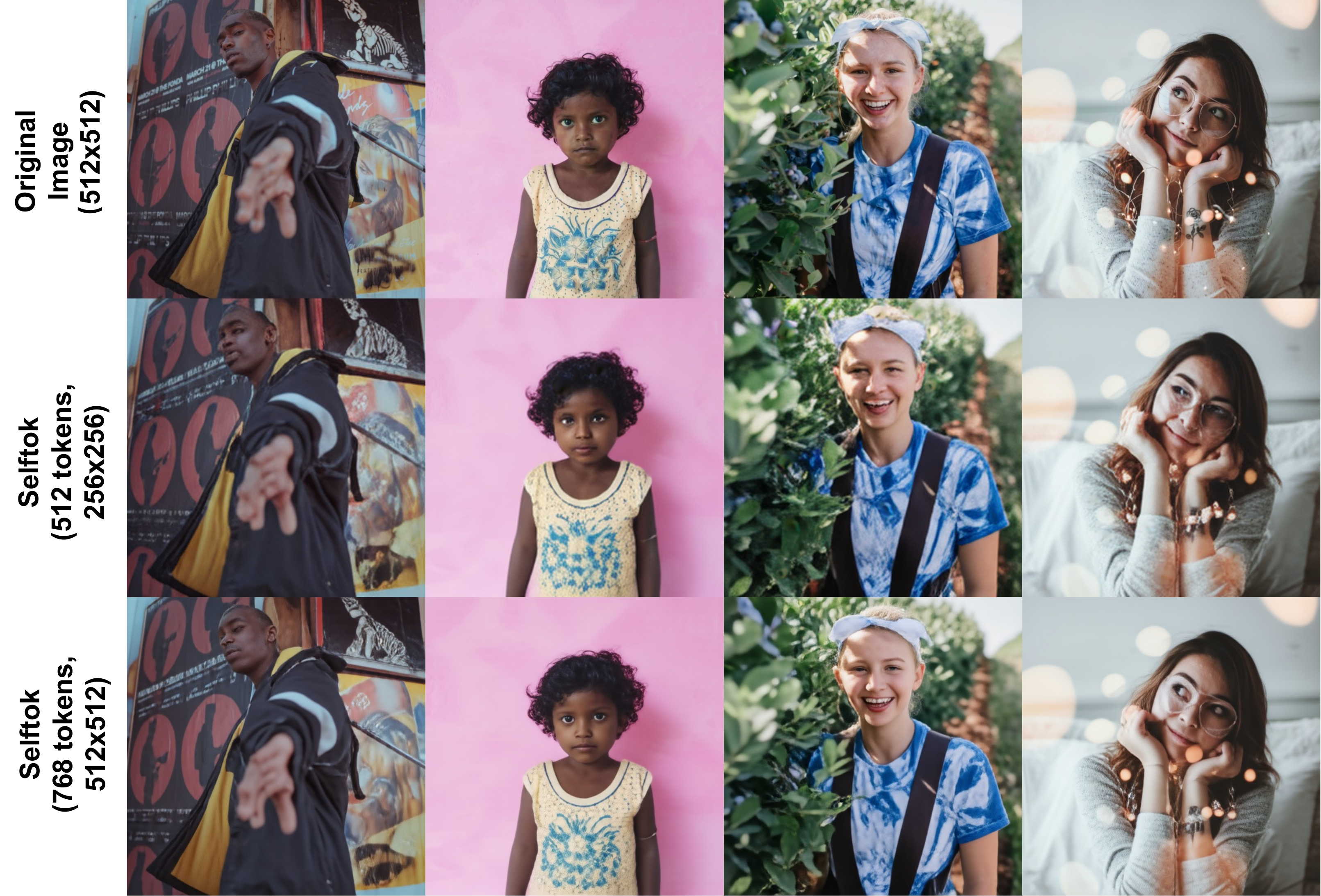}
    \vspace*{-5mm}
    \caption{Qualitative results of $512\times512$ resolutions.}
    \vspace*{-1mm}
    \label{fig:512}
\end{figure*}
\noindent\textbf{Multi-resolution Selftok}.
The current resolution of Selftok is limited to $256\times 256$, which constrains the quality of visual generation. Our design follows an incremental principle: higher-resolution images are supported by increasing the number of tokens, while reusing the tokens extracted from their lower-resolution counterparts. This enables efficient scalability, allowing higher-resolution data to leverage a dAR model pre-trained on lower-resolution inputs. This approach is particularly appealing, as it parallels the practice in LLM training, where longer document training benefits from prior training on shorter texts. Figure~\ref{fig:512} presents our preliminary results, which will be included in the next release of Selftok.

\noindent\textbf{Physics-aware Post-training}
Inspired by the impressive performance gains of visual RL by using the program-based reward, our next step is to incorporate physical laws into Selftok-based video generation. For example, we can track the trajectories of moving objects and evaluate whether they conform to fundamental motion principles. This direction has great potential in addressing the ever-lasting criticisms that large visual models struggle to learn a true world model~\cite{zhu2024sora,kang2024far}. In our recent work, we demonstrated that Selftok can achieve near-perfect object motion generation in a toy visual environment~\cite{lin2025phyvideo}.

\section{Contributions and Acknowledgments}

\definecolor{damaiblue}{RGB}{0, 0, 100}
\definecolor{damaigreen}{RGB}{0, 100, 0}
\definecolor{damaired}{RGB}{100, 0, 0}

\begin{multicols}{2} %

\noindent
\textbf{\color{damaired} Core Contributors} \\
\color{damaired} Bohan Wang \\
\color{damaired} Zhongqi Yue \\
\color{damaired} Fengda Zhang \\
\color{damaired} Shuo Chen \\
\color{damaired} Li'an Bi \\
\color{damaired} Junzhe Zhang \\
\color{damaired} Xue Song \\
\color{damaired} Kennard Yanting Chan \\
\color{damaired} Jiachun Pan \\
\color{damaired} Weijia Wu \\
\color{damaired} Mingze Zhou \\
\color{damaired} Wang Lin \\
\color{damaired} Kaihang Pan \\
\color{damaired} Saining Zhang \\
\color{damaired} Liyu Jia \\
\color{damaired} Wentao Hu \\

\noindent
\textbf{\color{damaired} Tech Lead} \\
\color{damaired} Hanwang Zhang \\

\columnbreak

\noindent
\textbf{\color{damaired} Project Lead} \\
\color{damaired} Wei Zhao \\

\noindent
\textbf{\color{damaiblue} Contributors} \\
\color{damaiblue} 
\color{damaiblue} Dongze Lian \\
\color{damaiblue} Zhaozheng Chen \\
\color{damaiblue} Wenzheng Xie \\
\color{damaiblue} Yiming Huang \\
\color{damaiblue} Yutong Hu \\
\color{damaiblue} Zeyi Huang \\
\color{damaiblue} Yuanlin Chen \\
\color{damaiblue} Shaokang Ma \\
\color{damaiblue} Mengchao Bai \\
\color{damaiblue} Wenhong Gong \\
\color{damaiblue} Lin Qi \\
\color{damaiblue} Lifei Zhu \\
\color{damaiblue} Tianjiao Guo \\

\end{multicols} %

\newpage
\appendix

\section*{Appendix}
\section*{A. VLM Training Data and Evaluation Details}

\subsection*{A.1 Training Data}

\begin{wrapfigure}{r}{0.4\textwidth}
\centering
\includegraphics[width=\linewidth]{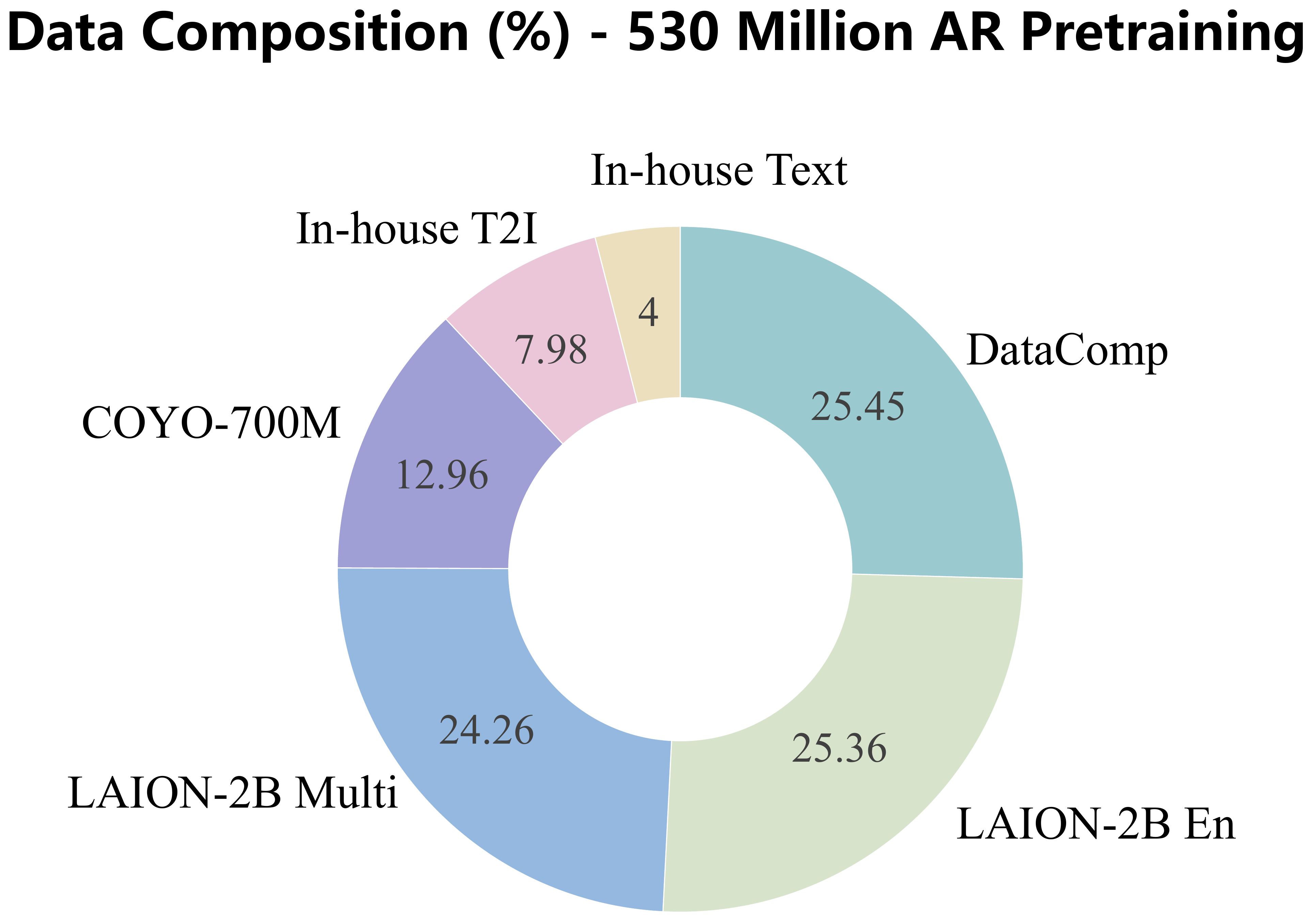}
\caption{\justifying The composition of 530 million AR pretraining data.}
\label{fig:ar_data}
\vspace{-3mm}
\end{wrapfigure}

\noindent\textbf{Cross-Modality Pretrain.} Our training data is comprised of both our in-house datasets and publicly available resources. These include datasets such as COYO~\cite{kakaobrain2022coyo-700m}, DataComp~\cite{datacomp}, and specific subsets of LAION-5B~\cite{schuhmann2022laion}. To ensure the high quality of the training data, we applied a series of rigorous filtering steps, including image resolution, NSFW, aesthetics, watermarking, etc. Together with our in-house text datasets, this results in a final high-quality dataset of 530 million image-text pairs and text sequences. A detailed breakdown of the dataset composition is provided in Figure \ref{fig:ar_data}. Along with the original captions available in existing datasets, we perform multi-level captioning~\cite{li2024playground} to capture various levels of information from the images. The multi-level captioning ranges from basic information to more complex elements, covering a spectrum from \textit{``subject > surrounding environment > subject action > visual language > camera language > world knowledge''}. We use the publicly available QwenVL~\cite{Qwen2-VL} and MiniCPM~\cite{hu2024minicpm} for multi-captioning (Example prompts used for multi-captioning are listed as follows). We perform training on 48 nodes with 384 Huawei Ascend 910B2 units. The global batch size is set to 7680, and the learning rate is set to 1e-4, with cosine decay to 1e-5. The total training steps are 200k iterations. We use the AdamW optimizer, with $\beta_1 = 0.9, \beta_2 = 0.95$. During training, one level of caption is randomly selected.

\begin{itemize}
    \item Level 0: “Summarize the contents of this image in no more than 5 words."
    \item Level 1: “Provide a concise caption (up to 10 words) that identifies the main subject of the image.”
    \item Level 2: “Create a short caption (up to 30 words) that describes the main subject of the image and includes a brief mention of its surrounding environment.”
    \item Level 3: “Generate a caption (up to 60 words) that describes the main subject, the surrounding environment. If the image is a photograph, include elements of visual language.”
    \item Level 4: “Create a medium-length caption that describes the main subject, its actions, the surrounding environment. If the image is a photograph, includes elements of visual language and camera language. Keep the caption around 50-100 words.”
    \item Level 5: “Create a caption that describes the main subject of the image, including its actions, the surrounding environment (from three aspects: foreground, mid-range, background), and any relevant world knowledge or named entities. If the image is a photograph, incorporate aspects of the image’s visual style (e.g., lighting, color tone, composition) and camera language (e.g., lens type, depth of field, angle) to enrich the description. Write the caption in fluent, natural language, and keep it around 150-200 words. Emphasize an engaging, informative tone that captures the essence of the scene. Consider any contextual clues from the setting to provide a well-rounded and insightful caption.”
\end{itemize}

\noindent\textbf{Supervised Fine-Tuning.} During this stage, we incorporate a variety of tasks for cross task training, outlined as follows: 

(1) Text-to-Image Generation: We curated tens of thousands of in-house image-text pairs for further improving the image-text alignment in the QT phase. In the training stage, the image resolution is set to 256 $\times$ 256. We perform training on one node with 8 Huawei Ascend 910B2 units. The global batch size is set to 32. There are 252,202 samples used in fine-tuning, constructed from our in-house image generation model. The short edges of all images were resized to 256 and then center-cropped to feed into the Selftok tokenizer. The learning rate is set to 1e-5. During inference, the logit adjustment scale is set to $10$ and the entropy threshold is $2.0$.

(2) Image Editing: The training data for image editing involves our in-house datasets as well as open-source ones, totaling 1,487,148 samples. These public datasets consist of Paint by InPaint Edit (PIPE)~\cite{wasserman2024paint}, UltraEdit~\cite{zhao2024ultraedit}, and OmniEdit~\cite{wei2024omniedit}, which are manually filtered to guarantee the quality, considering both the fidelity towards source images and the alignment between the instructions and corresponding image pairs.
Furthermore, we curated around two thousand in-house editing samples to further enhance the editing performance in the QT phase. The editing categories of our training data cover addition, removal, replacement, change attribution, and style transfer. In the training stage of image editing, the image resolution is set to 256 $\times$ 256. We perform training on two nodes with 8 Huawei Ascend 910B2 units each. The global batch size is set to 128. The large edges of all images are resized to 256, and then the small edges are white-padded to 256 before being fed into the Selftok tokenizer. The learning rate is set to 1e-5. During inference, the logit adjustment scale is set to $7.5$ and the entropy threshold is $2.0$.

\begin{table}[h]
\centering
\resizebox{\textwidth}{!}{
\fontsize{9}{11}\selectfont  
\begin{tabular}{p{3cm}|p{13cm}}
\toprule
\textbf{Dataset} & \textbf{Introduction} \\
\midrule
VQAv2 \cite{goyal2017vqav2}  & VQAv2, the updated version of the Visual Question Answering (VQA) dataset, consists of open-ended questions tied to images.
To answer these questions, models must have an understanding of visual content, language, and general world knowledge.
We transform the dataset into a dialogue format with the instruction: 'Answer the question using a single word or phrase.'\\

\midrule
OKVQA \cite{marino2019okvqa} & This dataset contains more than 14K questions that necessitate external knowledge for answering, with a focus on knowledge-driven visual question answering.  \\

\midrule
A-OKVQA \cite{schwenk2022aokvqa} & An enhanced version of OKVQA \cite{marino2019okvqa}, this dataset includes 25K questions that demand a wide range of commonsense and general knowledge for accurate answers. \\

\midrule
IconQA \cite{lu2021iconqa} & This dataset contains 107K questions divided into three sub-tasks, concentrating on abstract diagram interpretation and in-depth visual reasoning. \\

\midrule
GQA \cite{hudson2019gqa} & GQA is a comprehensive dataset comprising over 110K images and 22 million questions, designed to pair images with balanced question-answer sets for visual reasoning tasks.  \\

\midrule
OCR-VQA~\cite{mishra2019ocrvqa} & The OCR-VQA dataset is a dataset designed for visual question answering that combines optical character recognition (OCR), which contains 207,572 images of book covers and more than 1 million question-answer pairs about these images.
\\

\midrule
OneVision-Single Image~\cite{li2024llava} &
The LLaVA-OneVision dataset is a high-quality collection comprising 3.2 million images and over 7 million image-question pairs, spanning five major categories: General QA, General OCR, Document/Chart/Screen, Math Reasoning, and Language.\\

\midrule
TextVQA~\cite{singh2019towards} & The TextVQA dataset is a benchmark designed to advance VQA systems that require reading and reasoning about text within images. The training set contains 34,602 questions based on 21,953 images from OpenImages training set. \\

\midrule
PixMo~\cite{deitke2024molmo} & The PixMo dataset is a comprehensive collection designed to train the Molmo~\cite{deitke2024molmo} family of models, among which we use the PixMo-AskModelAnything and PixMo-CapQA subsets. AskModelAnything contains 162k human-authored question-answer pairs across 73k images, enabling models to respond to diverse image-related queries. CapQA features 214k synthetic question-answer pairs derived from 165k dense image captions. \\

\midrule 
VizWiz~\cite{gurari2018vizwiz} & VizWiz originates from real-world scenarios where blind users capture images using mobile phones and pose spoken questions about them. It contains 8k images to evaluate model’s zero-shot generalization on visual questions. \\

\bottomrule
\end{tabular}
}
\caption{Introduction of datasets used in Selftok's Vision Language Comprehension SFT. 
The formatting prompts outlined in \cite{liu2024improved} are used for processing non-dialogue datasets.
Note that only the training set is used for training.
}
\vspace*{-3mm}
\label{tab:mllm_data}
\end{table}

(3) Vision-Language Comprehension:
During the SFT stage, we used a variety of publicly available training data, as shown in Table~\ref{tab:mllm_data}.
Due to the significant differences in data volume across different datasets, we applied a proportional adjustment to the data sizes and then performed joint training.
In the training phase of vision-language comprehension, the image resolution is set to 256 $\times$ 256. 
We train on four nodes, each equipped with 8 Huawei Ascend 910B2 units, and the global batch size is set to 128.
The learning rate, similar to other SFT tasks, is set to 1e-5.
During inference, we standardize the format by using the instruction: 'Answer the question using a single word or phrase'.

\subsection*{A.2 Inference Process}
\begin{algorithm}[t]
  \small
  \caption{Adaptive Logit Adjustment Inference}
  \textbf{Input} model $\mathcal{M}$; conditional input $\mathbf{c}$; unconditional input $\mathbf{u}$; 
  
  initial generated tokens $\mathbf{r}$; entropy threshold $\tau$; adjustment scale $\gamma$
  \begin{algorithmic}[1]
    \State $\mathbf{r} \leftarrow \emptyset$
    \For{$i = 1, \dots, N$} \Comment{Iterate over token indices}
      \State $\mathbf{logit}_c \leftarrow \mathcal{M}(\mathbf{c}, \mathbf{r})$ \Comment{Compute logits for conditional input}
      \State $\mathbf{logit}_u \leftarrow \mathcal{M}(\mathbf{u}, \mathbf{r})$ \Comment{Compute logits for unconditional input}
      \If{$\text{entropy}(\mathbf{logit}_c) > \tau$} \Comment{Check if entropy exceeds threshold}
        \State $\mathbf{logit}_{\text{adjusted}} \leftarrow \mathbf{logit}_c + (\gamma - 1) \cdot (\mathbf{logit}_c - \mathbf{logit}_u)$ \Comment{Adjust logits based on entropy}
      \Else
        \State $\mathbf{logit}_{\text{adjusted}} \leftarrow \mathbf{logit}_c$ \Comment{No adjustment applied}
      \EndIf
      \State $\mathbf{logit}_{\text{top-k}} \leftarrow \text{top-k}(\mathbf{logit}_{\text{adjusted}})$ \Comment{Apply top-k sampling}
      \State $\mathbf{p}_{\text{top-k}} \leftarrow \text{softmax}(\mathbf{logit}_{\text{top-k}})$ \Comment{Normalize top-k logits to probabilities}
      \State $t_{\text{next}} \leftarrow \text{sample}(\mathbf{p}_{\text{top-k}})$ \Comment{Sample next token}
      \State $\mathbf{r} \leftarrow \mathbf{r} \cup t_{\text{next}}$ \Comment{Append sampled token}
    \EndFor
  \end{algorithmic}
  \textbf{Output} generated tokens $\mathbf{r}$
  \label{alg:adaptive-cfg}
\end{algorithm}
\begin{figure*}
    \centering
    \footnotesize
    \includegraphics[width=0.8\textwidth]{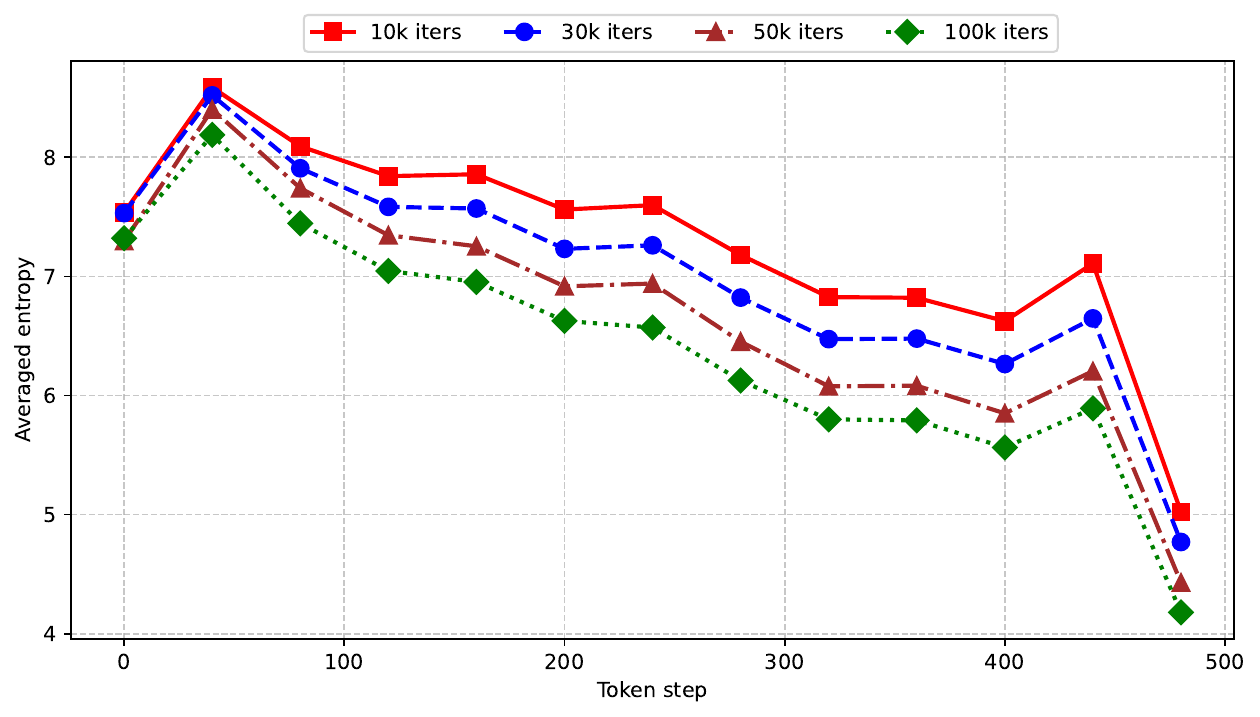}
    \vspace*{-5mm}
    \caption{Comparison of inference entropy at different iters.}
    \vspace*{-5mm}
    \label{fig:inference_entropy}
\end{figure*}
We propose an adaptive strategy for logit adjustment during inference, with the pseudo code presented in Algorithm~\ref{alg:adaptive-cfg}. Previous methods~\cite{chang2023muse, wang2024emu3} have referred to this logit adjustment as classifier-free guidance (CFG) . However, we note that the logit adjustment in AR models cannot be derived from Bayesian principles as done in diffusion-based approaches~(\cite{ho2022classifier,dhariwal2021diffusion}). Therefore, we prefer to refer to this process as logit adjustment, rather than CFG.

The core idea of our new approach is that logit adjustment is unnecessary when the token entropy is low. Low entropy indicates that the model has high confidence in its prediction for the current token, and further adjustment is not required. During the pretraining process, we also monitor entropy evolution. The Figure \ref{fig:inference_entropy} illustrates the change in token entropy across different timesteps during training. Key observations include: (1) as training progresses, the overall entropy decreases, suggesting that the model becomes more confident; and (2) as the number of generated tokens increases, the entropy for subsequent tokens decreases, indicating that the model is more certain in predicting future tokens.

It is important to highlight why language models in the AR paradigm do not require logit adjustment, whereas Selftok models do. While we believe that visual tokens in Selftok are similar to language tokens, insufficient training of visual tokens leads to high entropy at certain timesteps. Consequently, logit adjustment is needed to improve generation quality. However, with large-scale video token pretraining, Selftok models will overcome this issue, and similar to language models, logit adjustment will no longer be necessary during inference.

\subsection*{A.3 Evaluation Details}
\label{appendix:eval}

\textbf{Image Editing}: To quantitatively evaluate the editing capability, we employ the widely used PIE-Bench dataset \cite{ju2023direct} containing 700 testing samples with 10 types of editing: (0) random editing; (1) change object; (2) add object; (3) delete object; (4) change object content; (5) change object pose; (6) change object color; (7) change object material; (8) change image background; (9) change image style. For each editing type, source images are equally divided into artificial and natural ones, where each kind of image is further equally distributed to four scenes, \ie, animal, human, indoor, and outdoor. For metrics, we follow previous works \cite{ju2023direct,mu2025editar} in the perspective of both fidelity and editability using annotated foreground masks. Firstly, fidelity measures the similarity between source and edited images via the following metrics. Structure Distance ($\times 10^{3}$) calculates self-similarity matrices of DINO-ViT features and then applies cosine similarity between matrices of source and edited images. Apart from structure evaluation, PSNR, LPIPS ($\times 10^{3}$), MSE ($\times 10^{4}$), and SSIM ($\times 10^{2}$) are utilized to assess background preservation with annotated masks. Secondly, editability evaluates the alignment between the target image prompts and edited images. Specifically, the CLIP score ($\times 10^{2}$) is computed considering both the whole images and edited regions in the images. All metrics are compared at the resolution of $512 \times 512$.

\begin{figure*}
    \centering
    \footnotesize
    \includegraphics[width=0.9\textwidth]{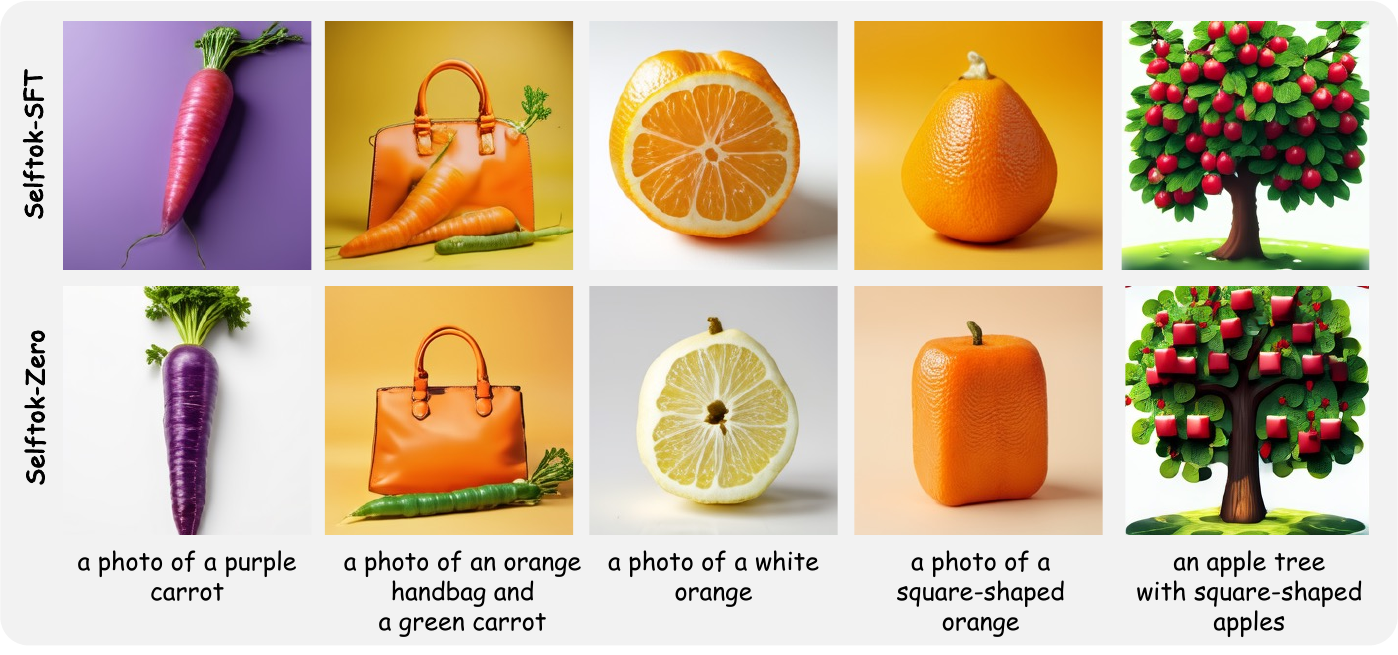}
    \vspace*{-3mm}
    \caption{More examples of failures in the Selftok-SFT model due to distributional biases in the training data during vision-language supervised training.}
    \vspace*{-3mm}
    \label{fig:app_hal}
\end{figure*}
\section*{B. Selftok-based Visual RL}
\label{sec:app_rl}
\noindent\textbf{Hallucination in Text-to-Image Generation.}
\label{sec:app_hal}
As discussed in the main text, one of the challenges in text-to-image generation is the “hallucination” issue, where a Vision-Language Model (VLM) tends to generate images that closely follow the training data distribution rather than genuinely reason about the text prompt. This can lead to the model failing to generate certain objects or scenes that are less common or not well-represented in the training set.
\begin{figure*}
    \centering
    \footnotesize
    \includegraphics[width=0.9\textwidth]{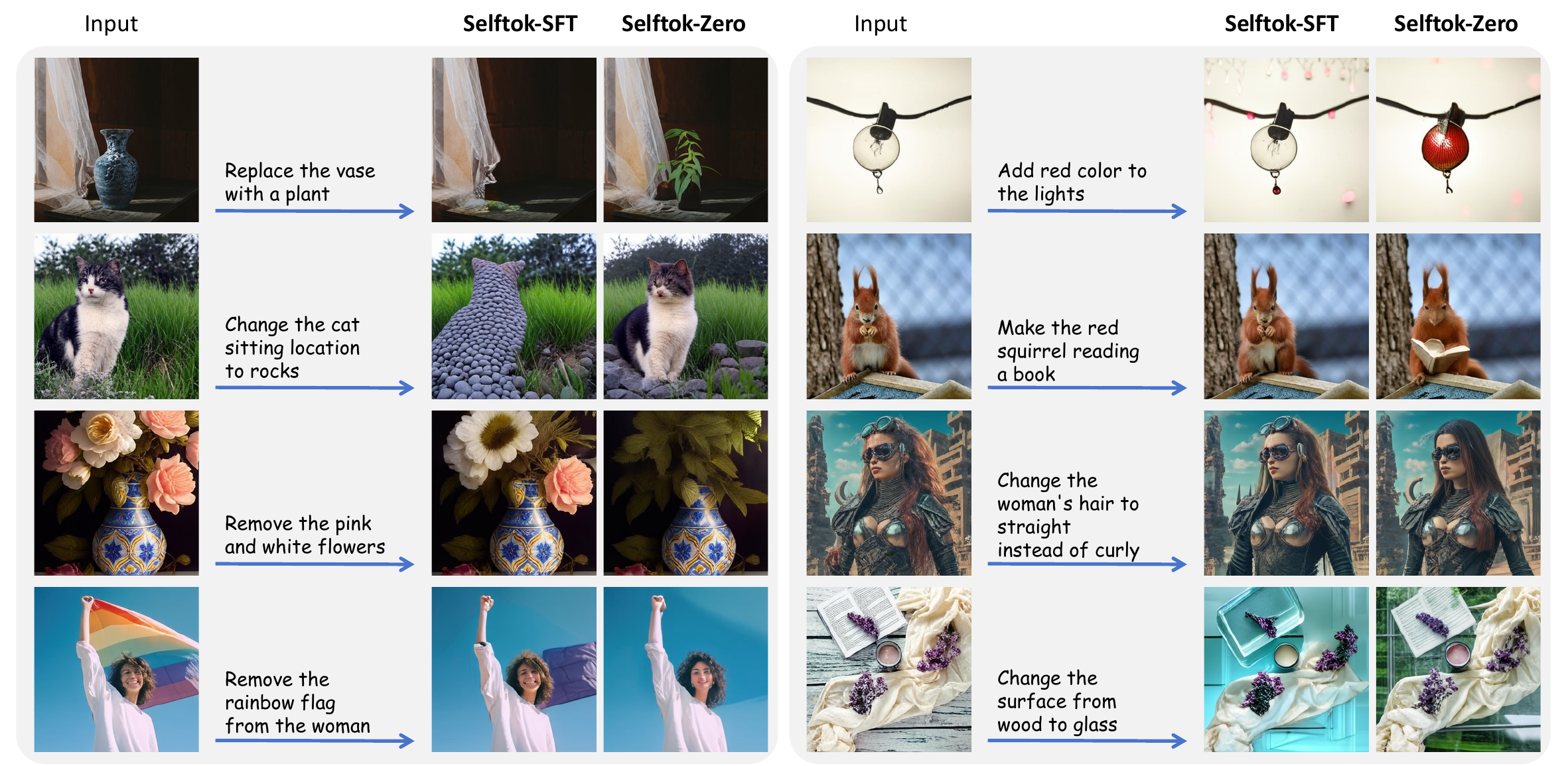}
    \vspace*{-3mm}
    \caption{Qualitative experimental results of Selftok-based visual RL on image editing. Compared to the Selftok-SFT, the images generated by Selftok-Zero demonstrate better alignment with the given instructions and better visual fidelity.}
    \vspace*{-3mm}
    \label{fig:app_edit}
\end{figure*}

In Figure~\ref{fig:app_hal}, we provide examples where the Selftok-SFT model fails to generate certain objects due to the rarity of these combinations in the training data. However, after applying visual RL (Selftok-Zero), the model is able to generate these previously missing combinations, showing a significant improvement in handling rare or complex prompts. The ability of Selftok-Zero to generate these images after the visual RL phase highlights how reinforcement learning can effectively overcome the hallucination problem, improving the model’s generalization and reasoning capabilities beyond the initial supervised training.

\noindent\textbf{Visual RL for Image Editing.}
\label{sec:app_rl_edit}
To further unlock the potential of our model, we also incorporate Visual RL into the image editing task, where we utilize a Vision-Language Model (VLM)—specifically, InternVL2.5-78B~\cite{internvl2.5}—as the reward model. This model evaluates whether the generated image strictly follows the instructions and accurately modifies the source image. Inspired by the work of~\cite{gu2024multi,gao2025seedream30technicalreport}, we ask the reward model to return a score between 0 and 5, with 5 indicating the highest level of adherence to the instructions. In a few hundred steps, our Selftok-Zero model shows a significant improvement over the Selftok-SFT model. As shown in Figure~\ref{fig:app_edit}, our model can correctly correspond to the instructions and generate appropriate edited images.

Unlike text-to-image generation, image editing involves more nuanced transformations, making it significantly more challenging to evaluate automatically.   The complexity arises from the need to assess both the fidelity of the edits to the original image and the accuracy of the applied changes according to the given instructions.   Therefore, to provide a more general and accurate reward for image editing tasks, we plan to explore more sophisticated reward models that can handle the intricacies of image modification.   Additionally, we aim to develop refined evaluation principles that can better capture the subtlety and precision required in image editing.   This will be a key focus in our future work, where we hope to improve the reliability of automated assessments and provide more meaningful feedback.

\setcounter{figure}{0}
\makeatletter 
\renewcommand{\thefigure}{A\@arabic\c@figure}
\makeatother

\setcounter{table}{0}
\makeatletter 
\renewcommand{\thetable}{A\@arabic\c@table}
\makeatother

\bibliographystyle{ieee_fullname}

{\scriptsize
\bibliography{main}
}

\end{document}